%% file: WWW_main.tex
\documentclass[sigconf]{acmart}
\usepackage{subcaption}
\usepackage{url}
\usepackage{amsmath, amsthm}
\usepackage{float}
\usepackage{algorithmic}
\usepackage{graphicx}
\usepackage{textcomp}
\usepackage{xcolor}
\usepackage{booktabs} 
\usepackage{multirow}
\usepackage{graphicx}
\usepackage{tcolorbox}

\AtBeginDocument{%
  }

\copyrightyear{2025}
\acmYear{2025}
\setcopyright{acmlicensed}
\acmConference[WWW '25] {Proceedings of the ACM Web Conference 2025}{April 28--May 2, 2025}{Sydney, NSW, Australia.}
\acmBooktitle{Proceedings of the ACM Web Conference 2025 (WWW '25), April 28--May 2, 2025, Sydney, NSW, Australia}
\acmISBN{979-8-4007-1274-6/25/04}
\acmDOI{10.1145/3696410.3714522}

\begin{document}

\title{Modality Interactive Mixture-of-Experts for Fake News Detection}


\author{Yifan Liu}
\affiliation{%
  \institution{School of Information Sciences}
  \institution{University of Illinois Urbana-Champaign}
  \city{Champaign}
  \state{Illinois}
  \country{USA}}
\email{yifan40@illinois.edu}

\author{Yaokun Liu}
\affiliation{%
  \institution{School of Information Sciences}
  \institution{University of Illinois Urbana-Champaign}
  \city{Champaign}
  \state{Illinois}
  \country{USA}}
\email{yaokunl2@illinois.edu}

\author{Zelin Li}
\affiliation{%
  \institution{School of Information Sciences}
  \institution{University of Illinois Urbana-Champaign}
  \city{Champaign}
  \state{Illinois}
  \country{USA}}
\email{zelin3@illinois.edu}

\author{Ruichen Yao}
\affiliation{%
  \institution{School of Information Sciences}
  \institution{University of Illinois Urbana-Champaign}
  \city{Champaign}
  \state{Illinois}
  \country{USA}}
\email{ryao8@illinois.edu}

\author{Yang Zhang}
\affiliation{%
  \institution{School of Information Sciences}
  \institution{University of Illinois Urbana-Champaign}
  \city{Champaign}
  \state{Illinois}
  \country{USA}}
\email{yzhangnd@illinois.edu}

\author{Dong Wang}
\affiliation{%
  \institution{School of Information Sciences}
  \institution{University of Illinois Urbana-Champaign}
  \city{Champaign}
  \state{Illinois}
  \country{USA}}
\email{dwang24@illinois.edu}




\input{abstract/abstract}




\maketitle
\renewcommand{\shortauthors}{Yifan Liu et al.}

\input{sections/introduction}
\input{sections/related_work}

\input{sections/methodology}
\input{sections/experiments}

\input{sections/discussion}

\section{Acknowledgments}
This research is supported in part by the National Science Foundation under Grant No. CNS-2427070,  IIS-2331069,  IIS-2202481, CHE-2105032, IIS-2130263, CNS-2131622. The views and conclusions contained in this document are those of the authors and should not be interpreted as representing the official policies, either expressed or implied, of the U.S. Government. The U.S. Government is authorized to reproduce and distribute reprints for Government purposes notwithstanding any copyright notation here on.

\clearpage
\bibliographystyle{ACM-Reference-Format}
\bibliography{reference}

\clearpage
\appendix
\input{sections/appendix}

\end{document}

%% file: abstract/abstract.tex
\begin{abstract}
The proliferation of fake news on social media platforms disproportionately impacts vulnerable populations, eroding trust, exacerbating inequality, and amplifying harmful narratives. Detecting fake news in multimodal contexts—where deceptive content combines text and images—is particularly challenging due to the nuanced interplay between modalities. Existing multimodal fake news detection methods often emphasize cross-modal consistency but ignore the complex interactions between text and visual elements, which may complement, contradict, or independently influence the predicted veracity of a post.
To address these challenges, we present \textbf{M}odality \textbf{I}nteractive \textbf{M}ixture-\textbf{o}f-\textbf{E}xperts for \textbf{F}ake \textbf{N}ews \textbf{D}etection (\textbf{MIMoE-FND}), a novel hierarchical Mixture-of-Expert framework designed to enhance multimodal fake news detection by explicitly modeling modality interactions through an interaction gating mechanism. Our approach models modality interactions by evaluating two key aspects of modality interactions: unimodal prediction agreement and semantic alignment. The hierarchical structure of MIMoE-FND allows for distinct learning pathways tailored to different fusion scenarios, adapting to the unique characteristics of each modality interaction. By tailoring fusion strategies to diverse modality interaction scenarios, MIMoE-FND provides a more robust and nuanced approach to multimodal fake news detection.
We evaluate our approach on three real-world benchmarks spanning two languages, demonstrating its superior performance compared to state-of-the-art methods. By enhancing the accuracy and interpretability of fake news detection, MIMoE-FND offers a promising tool to mitigate the spread of misinformation, with potential to better safeguard vulnerable communities against its harmful effects.
\end{abstract}

%% file: sections/introduction.tex
\section{Introduction}

\begin{figure*}[!htb]
    \centering
    \begin{subfigure}{0.23\textwidth}
        \begin{tcolorbox}[colframe=gray, colback=white, boxrule=0.5pt, left=2pt, right=2pt, top=2pt, bottom=2pt, boxsep=1pt, arc=0mm, outer arc=0mm]
        \parbox[t]{\textwidth}{\small \textbf{AM}: Semantically Misaligned; Unimodal Prediction Agreed. $(\hat{y}_{text}=0, \hat{y}_{img}=0, y=1)$}
        \includegraphics[width=\linewidth, height=80pt]{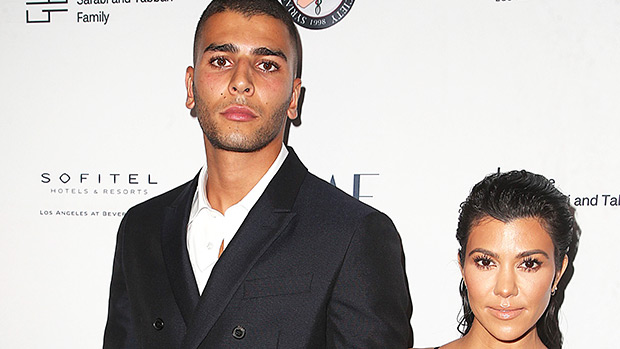}
        \parbox[t]{\textwidth}{\small HollywoodLife has now flip-flopped on its fake news stories that falsely maintained that John Cena and his fiancee Nikki Bella...}
        \end{tcolorbox}
        \caption{Agreed Misalignment}
        \label{fig:AM}
    \end{subfigure}
    \hfill
    \begin{subfigure}{0.23\textwidth}
        \begin{tcolorbox}[colframe=gray, colback=white, boxrule=0.5pt, left=2pt, right=2pt, top=2pt, bottom=2pt, boxsep=1pt, arc=0mm, outer arc=0mm]
        \parbox[t]{\textwidth}{\small \textbf{AA}: Semantically Aligned; Unimodal Prediction Agreed $(\hat{y}_{text}=1, \hat{y}_{img}=1, y=1)$}
        \includegraphics[width=\linewidth, height=80pt]{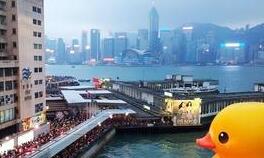}
        \parbox[t]{\textwidth}{\small While taking the luxury cruise ship passing the Big Yellow Duck, a tourist named Hu repeatedly threw 30 burning cigarette butts...}
        \end{tcolorbox}
        \caption{Agreed Alignment}
        \label{fig:AA}
    \end{subfigure}
    \hfill
    \begin{subfigure}{0.23\textwidth}
        \begin{tcolorbox}[colframe=gray, colback=white, boxrule=0.5pt, left=2pt, right=2pt, top=2pt, bottom=2pt, boxsep=1pt, arc=0mm, outer arc=0mm]
        \parbox[t]{\textwidth}{\small \textbf{DM}: Semantically Misaligned; Unimodal Prediction Disagreed $(\hat{y}_{text}=1, \hat{y}_{img}=0, y=1)$}
        \includegraphics[width=\linewidth, height=80pt]{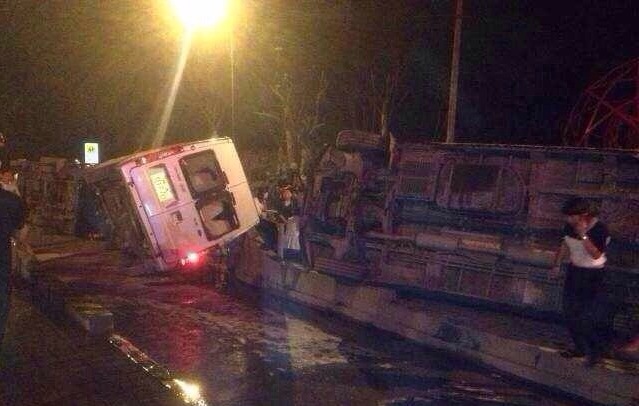}
        \parbox[t]{\textwidth}{\small The municipal government, without consulting the local residents, unilaterally decided to build the largest garbage incineration...}
        \end{tcolorbox}
        \caption{Disagreed Misalignment}
        \label{fig:DM}
    \end{subfigure}
    \hfill
    \begin{subfigure}{0.23\textwidth}
        \begin{tcolorbox}[colframe=gray, colback=white, boxrule=0.5pt, left=2pt, right=2pt, top=2pt, bottom=2pt, boxsep=1pt, arc=0mm, outer arc=0mm]
        \parbox[t]{\textwidth}{\small \textbf{DA}: Semantically Aligned; Unimodal Predictions Disagreed $(\hat{y}_{text}=0, \hat{y}_{img}=1, y=1)$}
        \includegraphics[width=\linewidth, height=80pt]{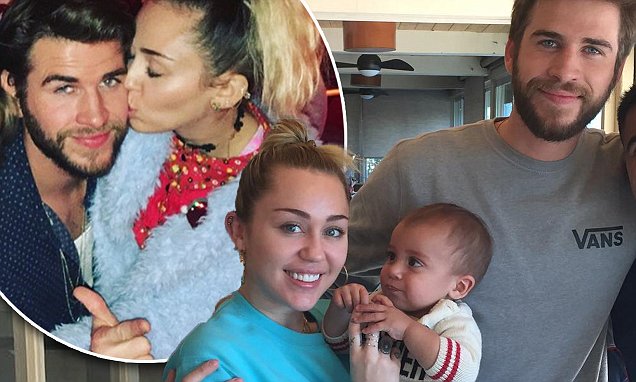}
        \parbox[t]{\textwidth}{\small Miley Cyrus and Liam Hemsworth 'to start a family in 2018' They met on the set of the 2009 film The Last Song, when they were...}
        \end{tcolorbox}
        \caption{Disagreed Alignment}
        \label{fig:DA}
    \end{subfigure}
    \caption{Examples of fake news with possible modality interactions. $\hat{y}_{text}$ and $\hat{y}_{img}$ denote the generated unimodal predictions, where positive class indicates fake news ($y=1$)}

    \label{fig:interaction example}
\end{figure*}
In recent years, the rise of online social networks has enabled users to freely express their opinions and emotions on the Web. However, this shift has also contributed to the proliferation of online fake news, defined as news content intentionally manipulated to spread disinformation or misinformation~\cite{shu2019fakenewsnetdatarepositorynews}. 
The rapid spread of fake news through social networks seriously threatens public knowledge and societal trust~\cite{spread, spread1},  
especially affecting vulnerable populations who are more susceptible to its influence~\cite{vulnerable_poplulation}. For example, alcohol-related deaths among young adults in the U.S. increased by 25\% in 2020~\cite{cahan2022alcohol}, fueled by COVID misinformation on social media that claimed consuming concentrated alcohol could kill the virus~\cite{islam2020covid}.

To combat this growing threat, automatic Fake News Detection (FND) has emerged as a critical research topic, aiming to develop solutions to safeguard online information integrity~\cite{FNDSurvey}. In FND research, textual cues (e.g., word choices and sentence sentiment~\cite{linguistic1, linguistic2, linguistic3}) and visual cues (e.g., image quality~\cite{quality} and image semantics~\cite{semAlignment}) provide complementary insights. To leverage both modalities for improved fake news detection, recent studies increasingly focus on multimodal FND ~\cite{FNDSurvey}.
Earlier works directly concatenate and fuse feature vectors from different modalities for classification~\cite{spotfake,spotfake+, EANN, zhou2023multimodal}.
However, these methods do not guide the model to prioritize specific unimodal representations or their shared cross-modal representations. For instance, in cases where the textual content is misleading but paired with an authentic image, the model might overly rely on the visual representation, leading to an incorrect prediction~\cite{modality_bias}. To address this challenge, prior work evaluates cross-modal consistency by leveraging similarity measures or incorporating auxiliary image-text matching tasks~\cite{CMC, CAFE, SAFE, coolant, bootstrap}.
Nevertheless, cross-modal consistency evaluation has limitations in fully capturing the nuanced interplay between modalities. First, existing cross-modal consistency evaluation methods
are under the assumption that real multimodal news (e.g., image-text pairs) are always semantically aligned~\cite{CMC, coolant, bootstrap}. Second, prior works ignore the agreement level of unimodal feature-only detections.

Specifically, we observe that in many online posts, images may not align or agree with the text content, yet the image-text pairs can still convey truthful or deceptive information, even when there are significant dissimilarities between the modalities. For instance, an image of a crowded hospital waiting room paired with text claiming a health crisis caused by vaccine side effects illustrates how mismatched visuals and text can craft a deceptive narrative to influence public opinion.
We refer to such scenarios in multimodal FND as \textbf{modality interactions}~\cite{paul_survey}. To the best of our knowledge, our work is the first to investigate modality interactions in the multimodal FND task while incorporating distinct modules to address the the variability of the interactions within the framework design.

In our approach, we focus on two key factors that define the interactions between modalities in a multimodal FND task: \textit{unimodal prediction agreement} and \textit{semantic alignment}. Specifically, \textit{unimodal prediction agreement} refers to the consistency between predictions of the veracity of the news (real or fake) based solely on unimodal features, while \textit{semantic alignment} captures the degree to which textual and visual content convey coherent or related meanings.

The combination of these two factors leads to four distinct types of modality interactions in FND, illustrated in Figure~\ref{fig:interaction example}:
1. \textbf{Agreed Misalignment (AM)}: semantically misalign, but unimodal predictions agree. For example, Figure~\ref{fig:AM} shows a celebrity image followed by a mismatched entertainment news. Despite the lack of semantic connection between the image and text, both unimodal predictions independently classify the instance as real news, emphasizing the challenge of extracting a meaningful relationship between unimodal predictions in scenarios where fake news can consist of both a true image and true text.
2. \textbf{Agreed Alignment (AA)}: semantically align, and unimodal predictions agree. In Figure~\ref{fig:AA}, the image and text both depict an incident of a large yellow duck toy, and their unimodal predictions both indicate fake news. This scenario highlights that similar signals across modalities can reinforce each other, generating a more confident detection.
3. \textbf{Disagreed Misalignment (DM)}: semantically misalign, and unimodal predictions disagree. Figure~\ref{fig:DM} shows a car crash image paired with unrelated text describing a new garbage incineration facility, with true image prediction and fake text prediction, highlighting the difficulty of reconciling contradictory signals from different modalities when they lack semantic overlap.
4. \textbf{Disagreed Alignment (DA)}: semantically align, but unimodal predictions disagree. As shown in Figure~\ref{fig:DA}, the text and image both indicate that Miley Cyrus and Liam Hemsworth starting a family, yet the image-only prediction estimates the post as fake news, while the text-only prediction identifies it as true news. This example reflects the challenge of resolving conflicts even with semantic alignment, introducing additional synergistic effects in semantic reasoning.

To address the limitations of prior approaches that overlook modality interactions (AM, AA, DM, DA), we propose \textbf{M}odality \textbf{I}nteractive \textbf{M}ixture-\textbf{o}f-\textbf{E}xperts for \textbf{F}ake \textbf{N}ews \textbf{D}etection (\textbf{MIMoE-FND}), a novel framework that explicitly models modality interactions through a gating mechanism, enabling tailored fusion strategies for improved detection accuracy. Overall, our method incorporates a hierarchical Mixture-of-Experts (MoE) architecture to dynamically route input data to different ``experts''~\cite{sparseMoE}. In our hierarchical approach, we first design improved MoE blocks for feature refinement and multimodal fusion through a token attention based gating, which enhance the model's ability to capture task-relevant information in different channels. At the upper level hierarchy of the MIMoE-FND architecture, we introduce a modality interaction gating module that dynamically routes input image-text pairs to distinct fusion expert modules. These fusion experts are trained to address different modality interaction challenges as discussed above, enabling a more tailored multimodal fusion for FND. To assess the efficacy of MIMoE-FND, we evaluate its performance on three real-world multimodal FND datasets in both English and Chinese, benchmarking it against state-of-the-art approaches. The results demonstrate significant improvements across four evaluation metrics, highlighting its superior accuracy and robustness. Our contributions are as follows:
\begin{itemize}
    \item We are the first to investigate and model modality interactions in the multimodal FND task. Our approach categorizes four distinct types of modality interactions based on semantic alignment and unimodal prediction agreement.
    \item We propose MIMoE-FND, a hierarchical MoE framework for multimodal FND task with adaptive multimodal fusion guided by modality interactions. Our model learns to dynamically route news instances to their corresponding fusion experts based on the evaluated modality interactions.
    \item We validate the effectiveness of MIMoE-FND through extensive experiments on three widely used multimodal FND benchmarks across two different languages, where our scheme shows significant performance gains compared to the state-of-the-art baselines.
\end{itemize}


%% file: sections/related_work.tex
\section{Related Work}
\subsection{Modality Interactions}
Modality interaction is a research topic that studies how elements in different modalities interact with each other to increase information for task inference~\cite{paul_survey}. For a multimodal downstream task, the effective task completion could require dynamic features extracted from both unimodal and multimodal inputs~\cite{paul_nonnegative}. To utilize information from different data modalities, factorized contrastive learning emerges as a framework accounting for different modality interactions in multimodal classification tasks with an information theoretical formulation~\cite{liang2023factorized, compressOrNot}. More recently, Yu et al. propose to model the modality interactions using a gating network~\cite{mmoe}. It highlights that different modality interactions could be better captured by separate modelings. In the realm of multimodal fake news detection, the detection model is expected to utilize both unique and shared information of different modalities to account for different modality interactions. However, existing multimodal fake news detection works do not explicitly model semantic-level modality interactions, resulting in suboptimal multimodal fusion. To this end, our work explicitly considers the modality interactions in multimodal FND through a gating mechanism supervised by a unimodal prediction agreement and semantic alignment. 

\subsection{Multimodal Fake News Detection}
Over the past few years, multimodal fake news detection has gained a significant amount of attention in research community~\cite{FNDSurvey}. Jin et al. proposes to utilize the attention mechanism to enhance LSTM model for effective modality fusion to detect online rumors~\cite{attRNN}. MVAE uses variational auto-encoders to learn a shared embedding space to account for both text and image data distribution in order to achieve a better multimodal fusion and more accurate FND~\cite{mvae}. SpotFake leverages pretrained XLNet and ResNet for feature extraction and perform FND based on the concatenated text-image representations
, benefiting from the rich features provided by these large pretrained models. However, despite the high-quality features these models offer for different data modalities, the misalignment of cross-modal features can reduce the FND performance.

To effectively align the text and visual representations in multimodal fake news detection, a number of prior works introduce extra modeling designs to account for modality-wise consistency to better guide the feature alignment heuristically~\cite{CAFE, SAFE, coolant}. SAFE models cross-modal inconsistency by calculating the similarity between text and visual information in news articles~\cite{SAFE}. MCAN uses co-attention layers to obtain fused feature from both visual and text inputs~\cite{MCAN}. MCNN models the modality-wise consistency by calculating the cosine similarity of visual and text representations after a weight sharing scheme~\cite{MCNN}. CAFE takes a probabilistic modeling approach by introducing two VAEs to model text and visual distributions, followed by a cross-modal ambiguity evaluation using Kullback-Leibler (KL) divergence~\cite{CAFE}. Similarly, CMC proposes to implicitly learn cross-modal correlation through knowledge distillation guided by a soft target~\cite{CMC}. COOLANT utilizes a cross-modal contrastive learning phase followed by a similar ambiguity-aware fusion as proposed in CAFE~\cite{coolant}. Likewise, with cross-modal ambiguity evaluation, BMR introduces a Mixture-of-Experts module to dynamically bootstrap multi-modal representations~\cite{bootstrap}. More recently, FND-CLIP uses pretrained CLIP feature representations to guide the fusion process of multi-modal FND, which brings a more semantic level cross-modal consistency evaluation~\cite{fnd_clip}.

The majority of prior work utilizing cross-modal ambiguity-guided fusion relies on weighted aggregation across modalities~\cite{CAFE, coolant, bootstrap}. While this method provides a mechanism to adaptively fuse different modality representations, it heavily focuses on statistical features due to the distribution modeling of cross-modal ambiguity. To this end, we propose to guide the fusion of multimodal representations using a gating mechanism supervised by both unimodal prediction agreement and CLIP-guided semantic alignment. Our method designs specialized fusion expert modules to account for different semantic/unimodal agreement scenarios, allowing flexibility for challenging cases in multimodal FND.

%% file: sections/methodology.tex
\begin{figure*}[htbp]
  \centering
      \includegraphics[width=0.8\textwidth]{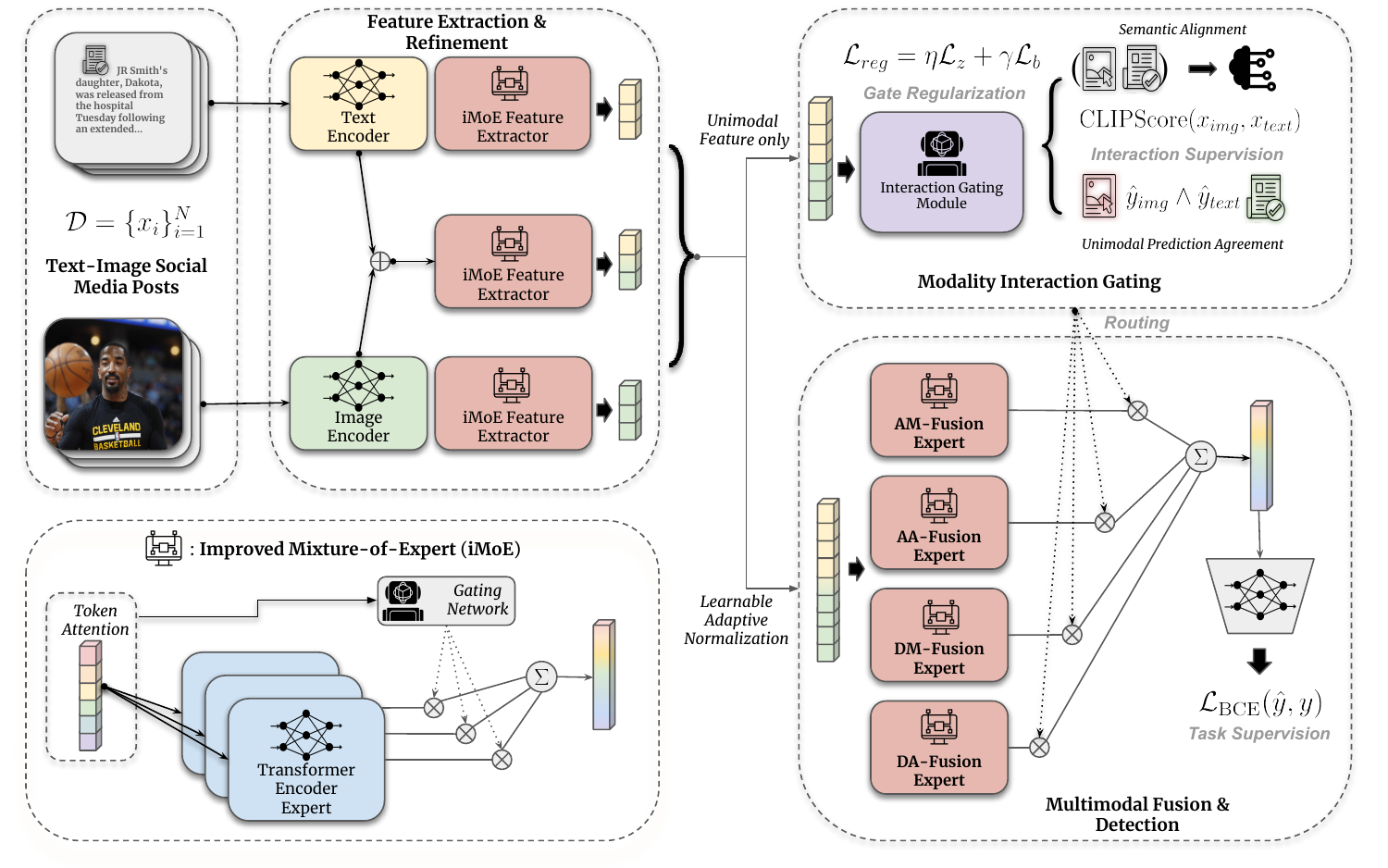}
  \caption{The pipeline of MIMoE-FND contains three phases: 1. \textbf{Feature Extraction \& Refinement}: BERT and MAE as unimodal encoders, followed by an iMoE module for feature refinement, 2. \textbf{Modality Interaction Gating}: a modality interaction gating network supervised by unimodal prediction agreement and CLIP-guided semantic alignment, 3. \textbf{Multimodal Fusion \& Detection}: four iMoE fusion experts to perform modality interaction gated fusion followed by a final classifier.}
  \label{fig:pipeline}
\end{figure*}

\vspace{-0.4cm}
\section{Method}

As shown in Figure~\ref{fig:pipeline}, our approach has a hierarchical MoE architecture, structured for three primary phases of our multimodal FND pipeline: 1. Feature Extraction \& Refinement, 2. Modality Interaction Gating, 3. Multimodal Fusion and Detection. In the lower-level hierarchy of MIMoE-FND, we utilize adapted MoE blocks for both feature refinement and multimodal feature fusion. These adapted MoEs are defined and elaborated as improved MoE blocks (iMoE) in Section~\ref{sec:imoe writeup}. For the upper-level hierarchy, we utilize a modality interaction gating network alongside distinct iMoE-based feature fusion experts to perform tailored multimodal fusion for samples based on their modality interactions (AM, AA, DM, DA). 

For the multimodal Fake News Detection (FND) task, we formulate it as a binary classification problem, where the positive class corresponds to \textit{fake news} and the negative class corresponds to \textit{real news}. The entire pipeline is trained in an end-to-end manner, taking inputs drawn from the dataset $\mathcal{D}$, where $\mathcal{D} = \{x_i \mid x_i = (x^{(i)}_{\text{text}}, x^{(i)}_{\text{img}})\}_{i=1}^N$. Here, each data point $x \in \mathcal{D}$ is a text-image pair, with $x_{\text{text}}$ representing the text input and $x_{\text{img}}$ representing the image input. The output of the pipeline is a binary detection result $\hat{y}$ indicating whether or not the input news is a fake news.

\subsection{iMoE Block}
\label{sec:imoe writeup}

We first outline our proposed improved Mixture-of-Experts block (iMoE) which is tailored to perform dynamic adaptive pooling along the token dimension of input. Similar to an MoE layer defined in~\cite{sparseMoE}, an iMoE block 
contains a gating network $G$ and $n$ expert networks $E_1, E_2, ..., E_n$. Inspired by the Squeeze-and-Excitation Network (SENet)~\cite{SENet}, we additionally introduce an attention vector, $att_x$, to enable the expert networks to flexibly focus on different tokens within an input vector $x$ consisting of $N$ tokens, each with $d$ dimensions.
Similar to an MoE layer, an iMoE block also consists of a gating network and $n$ expert networks. Taking inspiration from Squeeze-and-Excitation Network~\cite{SENet},
Specifically, as shown in Figure~\ref{fig:pipeline}, we compute $att_x$ by squeezing input $x$ along the token dimension using an attention network. The attention vector weighted input $(att_{x}\times x)$ is then passed to the gating network $G$ to obtain gate output, which we denote as a dispatch vector. Finally, the output $o$ of an iMoE block is given by a dispatch vector weighted feature aggregation from all expert networks: $o = \sum_i G_i(att_{x} \times x) E_i(x)$.

In the iMoE block, both the attention network and the gating network are composed of two fully connected layers, with a Sigmoid activation function applied between them. For each expert network $E$ within the iMoE, we utilize a pre-defined transformer block derived from the Vision Transformer~\cite{vit}.


\subsection{Feature Extraction \& Refinement}
In the feature extraction \& refinement phase of MIMoE-FND pipeline, we utilize pretrained BERT~\cite{bert} to extract text representations and MAE~\cite{MAE} to extract image representations for a given input $x = (x_{\text{text}}, x_{\text{img}})$. More specifically, we denote the extracted unimodal features as $u_t \in \mathbb{R}^{N_t \times D_{BERT}}$ for text and $u_i \in \mathbb{R}^{N_i \times D_{MAE}}$ for image. 

While BERT and MAE provide rich semantic features, directly combining image and text representations can be challenging due to the discrepancies between text and image pretrained feature spaces. To mitigate this issue, we use iMoE blocks to refine the unimodal representations and perform adaptive pooling. We construct three feature branches for unimodal text representation $u_t$, unimodal image representation $u_i$ and multimodal representation $u_{m} = concat(u_t, u_i)$. For each feature branch, we pass the feature vector to an iMoE block to obtain the pooled and refined feature vector. We denote the refined feature vectors as $e_t, e_i, e_m$ for unimodal text, unimodal image and multimodal representations respectively. 


\subsection{Modality Interaction Gating}
To route news instances with different modality interactions to their respective feature fusion experts, we introduce a modality interaction gating mechanism. The modality interaction gating module is supervised by evaluations of \textit{unimodal prediction agreement} and \textit{semantic alignment}, ensuring that the gating process effectively leverages modality interactions.
\subsubsection{Modality Interactions} 
To quantitatively evaluate the modality interaction of an instance, we define unimodal prediction divergence and semantic alignment as follows:

\begin{definition}[Unimodal Prediction Divergence]
For a given input news $x$, we define unimodal prediction agreement to be: 
{\small
\begin{equation}
\delta(x) = d(\hat{y}_{\text{text}}, \hat{y}_{\text{img}})
\end{equation}}
where $\hat{y}_j$ represents the fake news detection result of the news instance based solely on the information from modality $j$ (e.g., text or image), and $d: \mathcal{Y} \times \mathcal{Y} \rightarrow \mathbb{R}^+$ is a distance function measuring the divergence between unimodal predictions.
\end{definition}

\begin{definition}[Semantic Alignment]
For a given input news $x$, we define the semantic alignment as the semantic level similarity between the text and image. Let $\alpha_{text}$ and $\alpha_{img}$ be two vector representations fully capturing semantic features in a shared feature space of text and image. The semantic alignment is defined as: 
{\small \begin{equation}
\label{eq:semantic_alignment}
\rho(x) = \frac{\alpha_{\text{text}} \cdot \alpha_{\text{img}}}{\|\alpha_{\text{text}}\| \|\alpha_{\text{img}}\|}
\end{equation}}
\end{definition}

To evaluate the unimodal prediction divergence $\delta(x)$, we attach classification heads $f_{\text{text}}, f_{\text{img}}$ for each of the refined unimodal feature branch to extract $\hat{y}_\text{text} = f_{\text{text}} (e_t)$ and $\hat{y}_\text{img} = f_{\text{img}}(e_i)$. The classification heads are trained as a separate part of the pipeline, which is not used in our final prediction. We use the standard cross-entropy loss to guide the unimodal classification heads: {\small \begin{equation}
\mathcal{L}_{uni}(\hat{y}_{text}, \hat{y}_{img}) = \frac{1}{2}(\mathcal{L}_{CE}(\hat{y}_{text}, y) + \mathcal{L}_{CE}(\hat{y}_{img}, y))
\end{equation}}

We then evaluate $\delta(x)$ to be the Jensen-Shannon (JS) divergence~\cite{MENENDEZ1997307} between unimodal predictions, defined as:
{\small
\begin{equation}
\delta(\mathbf{x}) = D_{JS}(\hat{y}_{\text{text}}, \hat{y}_{\text{img}}) = \frac{1}{2} D_{KL}(\hat{y}_{\text{text}} \| M) + \frac{1}{2} D_{KL}(\hat{y}_{\text{img}} \| M), 
\end{equation}}
where $M = \frac{1}{2} (\hat{y}_{\text{text}} + \hat{y}_{\text{img}})$ represents the mean of the unimodal prediction distributions, $D_{KL}$ denotes the Kullback-Leibler (KL) divergence and $\hat{y}_{j}$ is the unimodal prediction distribution of modality $j$ (e.g., text or image).

For the evaluation of semantic alignment, we pass the raw text-image pair to a pretrained CLIP model to extract feature vectors $m_{t}$ and $m_{i}$ for text and image in a joint feature space. CLIP is pretrained on a variety of internet-sourced text-image pairs, which enables them to capture semantic relationship between texts and images. The calculation of $\rho(x)$ in equation~\ref{eq:semantic_alignment} is thus computed as the cosine similarity of CLIP text and image embeddings, which is also referred to as the CLIP score~\cite{clipScore}.

Additionally, we define an agreement threshold $\theta_{agr}$ and an alignment threshold $\theta_{sem}$, which are set to fixed constants in all our experiments empirically. Specifically, we categorize a data instance to be \textit{semantically aligned} and \textit{unimodal prediction agreed} based on $\mathbf{1}(\rho(x) > \theta_{sem})$ and $\mathbf{1}(\delta(x) < \theta_{agr})$ respectively, where $\mathbf{1}(\cdot)$ is an indicator function with binary outputs (0 or 1). We categorize four modality interactions (AM, AA, DM, DA) in a multimodal FND task according to the permutations of outputs of the binary indicators.  

\subsubsection{Interaction Gating Module}
We introduce an interaction gating module that learns to dispatch news instances according to their modality interactions while also optimizing task performance. The interaction gating module takes the refined unimodal feature vectors $(e_{t}, e_{i})$ and the CLIP embeddings $(m_{t}, m_{i})$ as inputs, and outputs a dispatch vector $\hat{y}_{d}$ with a size corresponds to the number of feature fusion experts. Similar to the gating network illustrated in the iMoE block (section~\ref{sec:imoe writeup}), the interaction gating module contains an attention network and an interaction gating network. The attention network is used to calculate a modality-level attention vector. With the modality-level attention weighted input, the interaction gating module computes a softmax-normalized dispatch vector which determines the routing process. During the modality interaction routing, only the fusion expert corresponding to the highest value in the dispatch vector is activated, ensuring that multimodal features are processed by the most relevant fusion expert.

To guide the training of interaction gating network to route the data instances according to modality interactions, we use $\mathbf{1}(\rho(x) > \theta_{sem})$ and $\mathbf{1}(\delta(x) < \theta_{agr})$ as our training target for interaction gating by expanding the two binary outputs to 4 different classes corresponding to modality interactions (AM, AA, DM, DA), where we obtain the target modality interaction $y_{\text{int}} = 2 \cdot \mathbf{1}(\delta(x) < \theta_{agr}) + \mathbf{1}(\rho(x) > \theta_{sem})$. The modality interaction gating process is then formulated as a classification task, where the model output is $\hat{y}_{d}$ and the target label is $y_{int}$. Given a dispatch vector output $\hat{y}_{d}$ from the gating network, we define a modality interaction loss as follows: 
{\small
\begin{equation}
\begin{aligned}
    \mathcal{L}_{d} = \mathcal{L}_{CE}(\hat{y}_{d}, y_{int})
\end{aligned}
\end{equation}
}
\begin{table*}[t]
\centering
\caption{Performance Metrics for Different Classification Methods on Weibo, GossipCop and Weibo-21}
\vspace{-0.2cm}
\label{tab:performance}
\resizebox{0.8\textwidth}{!}{
\begin{tabular}{@{}clcccccccc@{}}
\toprule
\textbf{Dataset} & \textbf{Method} & \textbf{Accuracy} & \multicolumn{3}{c}{\textbf{Fake News}} & \multicolumn{3}{c}{\textbf{Real News}} \\
\cmidrule(lr){4-6} \cmidrule(lr){7-9}
& & & \textbf{Precision} & \textbf{Recall} & \textbf{F1-score} & \textbf{Precision} & \textbf{Recall} & \textbf{F1-score} \\
\midrule
\multirow{5}{*}{Weibo}     & EANN~\cite{EANN} & 0.827 & 0.847 & 0.812 & 0.829 & 0.807 & 0.843 & 0.825 \\
                           & SAFE~\cite{SAFE} & 0.762 & 0.831 & 0.724 & 0.774 & 0.695 & 0.811 & 0.748 \\
                           & SpotFake~\cite{spotfake} & 0.892 & 0.902 & \textbf{0.964} & \textbf{0.932} & 0.847 & 0.656 & 0.739 \\
                           & CAFE~\cite{CAFE} & 0.840 & 0.855 & 0.830 & 0.842 & 0.825 & 0.851 & 0.837 \\
                           & BMR~\cite{bootstrap} & 0.918 & 0.882 & 0.948 & 0.914 & \textbf{0.942} & 0.879 & 0.904 \\
                           & FND-CLIP~\cite{fnd_clip} & 0.907 & 0.914 & 0.901 & 0.908 & 0.914 & 0.901 & 0.907 \\
                           & \textbf{MIMoE-FND} & \textbf{0.928} & \textbf{0.942} & 0.913 & 0.928 & 0.913 & \textbf{0.942} & \textbf{0.927} \\
\midrule
\multirow{5}{*}{GossipCop} & EANN~\cite{EANN} & 0.864 & 0.702 & 0.518 & 0.594 & 0.887 & 0.956 & 0.920 \\
                           & SAFE~\cite{SAFE} & 0.838 & 0.758 & 0.558 & 0.643 & 0.857 & 0.937 & 0.895 \\
                           & SpotFake~\cite{spotfake} & 0.858 & 0.732 & 0.372 & 0.494 & 0.866 & 0.962 & 0.914 \\
                           & CAFE~\cite{CAFE} & 0.867 & 0.732 & 0.490 & 0.587 & 0.887 & 0.957 & 0.921 \\
                           & BMR~\cite{bootstrap} & \textbf{0.895} & 0.752 & 0.639 & 0.691 & \textbf{0.920} & \textbf{0.965} & \textbf{0.936} \\
                           & FND-CLIP~\cite{fnd_clip} & 0.880 & 0.761 & 0.549 & 0.638 & 0.899 & 0.959 & 0.928 \\
                           & \textbf{MIMoE-FND} & \textbf{0.895} & \textbf{0.762} & \textbf{0.644} & \textbf{0.698} & \textbf{0.920} & 0.953 & \textbf{0.936} \\
 

\midrule
\multirow{5}{*}{Weibo-21}  & EANN~\cite{EANN} & 0.870 & 0.902 & 0.825 & 0.862 & 0.841 & 0.912 & 0.875 \\
                           & SAFE~\cite{SAFE} & 0.905 & 0.893 & 0.908 & 0.901 & 0.916 & 0.901 & 0.890 \\
                           & SpotFake~\cite{spotfake} & 0.851 & 0.953 & 0.733 & 0.828 & 0.786 & \textbf{0.964} & 0.866 \\
                           & CAFE~\cite{CAFE} & 0.882 & 0.857 & 0.915 & 0.885 & 0.907 & 0.844 & 0.876 \\
                           & BMR~\cite{bootstrap} & 0.929 & 0.908 & 0.947 & 0.927 & 0.946 & 0.906 & 0.925 \\
                           & FND-CLIP~\cite{fnd_clip} & 0.943 & 0.935 & 0.945 & 0.940 & 0.950 & 0.942 & 0.946 \\                           
                           & \textbf{MIMoE-FND} & \textbf{0.956} & \textbf{0.953} & \textbf{0.957} & \textbf{0.955} & \textbf{0.959} & 0.956 & \textbf{0.957} \\
\bottomrule 
\end{tabular}
}
\end{table*}

\vspace{-0.5cm}
\subsubsection{Gating Regularization}

In our experiments, we observe that the modality interaction gating mechanism can lead to imbalanced training of the feature fusion experts, causing suboptimal performance in the final detection task. We hypothesize that this issue arises from an imbalance in the training data distribution across different modality interactions. To ensure effective training of interaction gating modules and the associated fusion experts, we apply a router-Z loss to penalize extreme values in interaction gating network outputs and a balance loss to ensure load balancing of fusion experts~\cite{Zoph2022STMoEDS}. For a dispatch vector $\hat{y}_{d}$, the router-Z loss $\mathcal{L}_{z}$ and the balance loss $\mathcal{L}_{b}$ are calculated as following:
{\small
\begin{equation}
\begin{aligned}
    \mathcal{L}_{z} = (\log \sum \exp(o_d^{(i)}))^2, \quad \mathcal{L}_{b} = \frac{1}{N} \sum (\hat{y}_{d}^{(i)} - t^{(i)}) 
\end{aligned}
\end{equation}
}
where $\hat{y}_d$ denotes the softmax normalized dispatch vector, $o_d$ denotes the raw output of interaction gating network and $t^{(i)}$ denotes the probability from a uniform distribution $t$ of data instance being assigned to expert $i$. Overall, apart from the task loss of multimodal FND, the interaction gating module is supervised by a combination of modality interaction loss $\mathcal{L}_d$ and regularization loss $\mathcal{L}_{reg} = \eta \mathcal{L}_{z} + \gamma \mathcal{L}_{b}$:
{\small
\begin{equation}
    \mathcal{L}_{int} = \mathcal{L}_{d} + \mathcal{L}_{reg}
\end{equation}}
where we empirically set the regularization weights $\eta=0.01$ and $\gamma=0.1$ respectively in our experiments.

\subsection{Multimodal Fusion \& Detection}
\label{sec:experts}
In the multimodal fusion and detection phase, we first apply adaptive normalization that adjusts the unimodal feature vectors $e_t, e_i$ with learnable mean and variance~\cite{adain} to further reduce the feature space discrepancy for better fusion, where we denote the normalized unimodal features as $e'_t$ and $e'_i$. For each data instance, we then perform fusion by passing concatenated the adaptively normalized feature vectors $(e'_t, e_m, e'_i)$ to the dispatched fusion expert. 

In MIMoE-FND, we use four fusion experts corresponding to modality interactions (AM, AA, DM, DA). Each fusion expert is an iMoE block performing modality-level aggregation. The final feature vector is obtained from the target fusion expert guided dispatch vector $\hat{y}_d$. In the detection phase, the final feature vector is fed into a classification head to obtain a detection result $\hat{y}$, which is then used to calculate the task loss as binary cross-entropy loss.
Our overall objective function is formulated as a weighted sum of modality interaction gating module loss, task loss and unimodal prediction loss, which can be summarized as:
{\small
\begin{equation}
    \mathcal{L} = \mathcal{L}_{task} + \alpha \mathcal{L}_{uni} + \beta \mathcal{L}_{int}
\end{equation}}
where the $\alpha$ and $\beta$ are hyper-parameters adjusting the different learning speeds for the components. The whole pipeline is then trained in an end-to-end manner with a separate optimizer to update parameters for the interaction gating module and an optimizer for all other parameter updates.

%% file: sections/experiments.tex
\section{Experiments}

To evaluate our approach, we benchmark it against several strong multimodal FND baselines on three widely used datasets: Weibo~\cite{weibo}, Weibo-21~\cite{weibo21}, and GossipCop~\cite{gossipcop}. The detailed experimental settings and the list of baseline methods are provided in Appendix~\ref{experimental_setting}.

\vspace{0cm}
\subsection{Performance Comparison}
The performance comparisons between our approach and other baselines are reported in Table~\ref{tab:performance}. We report accuracy, precision, recall and F1-score. We observe that our method is able to consistently perform the best in terms of detection accuracy across all datasets with the accuracy of 0.928, 0.895 and 0.956 for Weibo, GossipCop and Weibo-21 respectively. Our method is also able to out-perform most of the baselines on other reported metrics for both the fake news and real news. However, we do observe several outliers in the results. For Weibo, we notice that SpotFake has better fake news recall and F1-score. However, it has the worst performance for the real news prediction. Such a severely imbalanced prediction indicates that the model has not been generalized well to the data distribution and is instead overfitting to the characteristics of fake news, failing to correctly identify real news. 

For GossipCop, BMR has a very close performance compared to our approach. Nevertheless, we notice our method is able to perform better in all metrics for fake news detection while tie with BMR in terms of real news precision and F1-score. This suggests that our method can detect fake news more effectively while keeping the same real news prediction performance as BMR. To further understand the performance of our approach, we closely scrutinize GossipCop dataset and find the majority data collected from GossipCop are well-edited entertainment news with celebrity faces as the accompanied image, which can only partially reflect the modality interactions. On the other hand, Weibo and Weibo-21 contain more noisy user-generated posts, thus reflecting a more complex set of modality interactions, where we observe that our method is able to obtain larger performance gains.


\vspace{-0.2cm}
\subsection{Ablation Study}
To understand the effectiveness of each components in our approach, we perform ablation study on input modalities and modality interaction gating as shown in Table~\ref{tab:ablation} with following settings:

\vspace{-5pt}
\begin{itemize}
    \item \textbf{image-only}: We use only image input followed by a feature refinement component and an MLP classification head.
    \item \textbf{text-only}: We use only text followed by the same feature refinement component and classification head.
    \item \textbf{w/o reg.}: We remove the regularization losses $\mathcal{L}_z$ and $\mathcal{L}_b$ in the loss function of interaction gating network while keep all other components and hyper-parameters the same.
    \item \textbf{w/o sem.}: We remove the semantic alignment (sem.) target in the training of interaction gating network. 
    \item \textbf{w/o agr.}: We remove the unimodal prediction agreement (agr.) target in the training of interaction gating network.
    \item \textbf{w/o int.}: We remove the modality interaction (int.) supervision (both sem. and agr.) and use final task loss together with regularization losses to guide interaction gating network.  
\end{itemize}
\vspace{-5pt}

\begin{table}[t]
\caption{Ablation Study}
\vspace{-0.3cm}
\label{tab:ablation}
\resizebox{0.45\textwidth}{!}{
\begin{tabular}{@{}clcccccccc@{}}
\toprule
\textbf{Dataset} & \textbf{Method} & \textbf{Accuracy} & \textbf{F1(Fake)} & \textbf{F1(Real)} \\
\midrule
\multirow{5}{*}{Weibo}     & image only & 0.818 & 0.825 & 0.811 \\
                           & text only & 0.888 & 0.774 & 0.893 \\
                           & w/o reg. & 0.827 & 0.830 & 0.825 \\
                           & w/o sem. & 0.815 & 0.817 & 0.812 \\
                           & w/o agr. & 0.815 & 0.820 & 0.810 \\
                           & w/o int. & 0.916 & 0.915 & 0.916 \\
                           & \textbf{MIMoE-FND} & \textbf{0.928} & \textbf{0.928} & \textbf{0.927} \\

\midrule
\multirow{5}{*}{GossipCop} & image only & 0.820 & 0.363 & 0.895 \\
                           & text only & 0.882 & 0.670 & 0.928 \\
                           & w/o reg. & 0.818 & 0.067 & 0.899 \\
                           & w/o sem. & 0.824 & 0.239 & 0.900 \\                          
                           & w/o agr. & 0.823 & 0.282 & 0.898 \\                   
                           & w/o int. & 0.887 & 0.665 & 0.932 \\           
                           & \textbf{MIMoE-FND} & \textbf{0.895} & \textbf{0.698} &\textbf{0.936} \\

\midrule
\multirow{5}{*}{Weibo-21}  & image only & 0.843 & 0.830 & 0.853 \\
                           & text only & 0.925 & 0.922 & 0.928 \\
                           & w/o reg. & 0.860 & 0.847 & 0.872 \\
                           & w/o sem. & 0.865 & 0.864 & 0.867 \\
                           & w/o agr. & 0.862 & 0.859 & 0.865 \\
                           & w/o int. & 0.922 & 0.921 & 0.922 \\
                           & \textbf{MIMoE-FND} & \textbf{0.956} & \textbf{0.955} & \textbf{0.957} \\
\bottomrule 
\end{tabular}}
\end{table}
\setlength{\textfloatsep}{5pt}

We use the same training procedure as our main experiments to better understand the contributions of modalities and network components. For unimodal prediction performance, we observe that text-only based prediction consistently achieves better accuracy than image, indicating text contains more task-relevant information for multimodal FND. We observe that our method can consistently achieve more robust and accurate FND by a significant margin compared to unimodal baselines. 

For different components of modality interaction gating module, we observe that the removal of the regularization losses result in a significant accuracy decline. Specifically, the accuracy fell to between the image-only and text-only prediction accuracies in evaluations using the Chinese datasets. For the GossipCop dataset, the accuracy deteriorated further, performing worse than both unimodal predictions. This observation underscores the importance of the introduced regularization—balance loss and router-Z loss—in our training process, as these components are crucial for precise input dispatch and load balancing among fusion experts \cite{Zoph2022STMoEDS}. Additionally, we hypothesize that the poorer performance on GossipCop is attributable to the more imbalanced occurrence of modality interactions within this dataset, which results in insufficient training of some fusion experts without regularization.
\begin{figure}[t]
    \setlength{\abovecaptionskip}{0pt}
    \setlength{\belowcaptionskip}{3pt}
    \centering
    \begin{subfigure}{\columnwidth}
    \centering
    \includegraphics[width=\columnwidth]{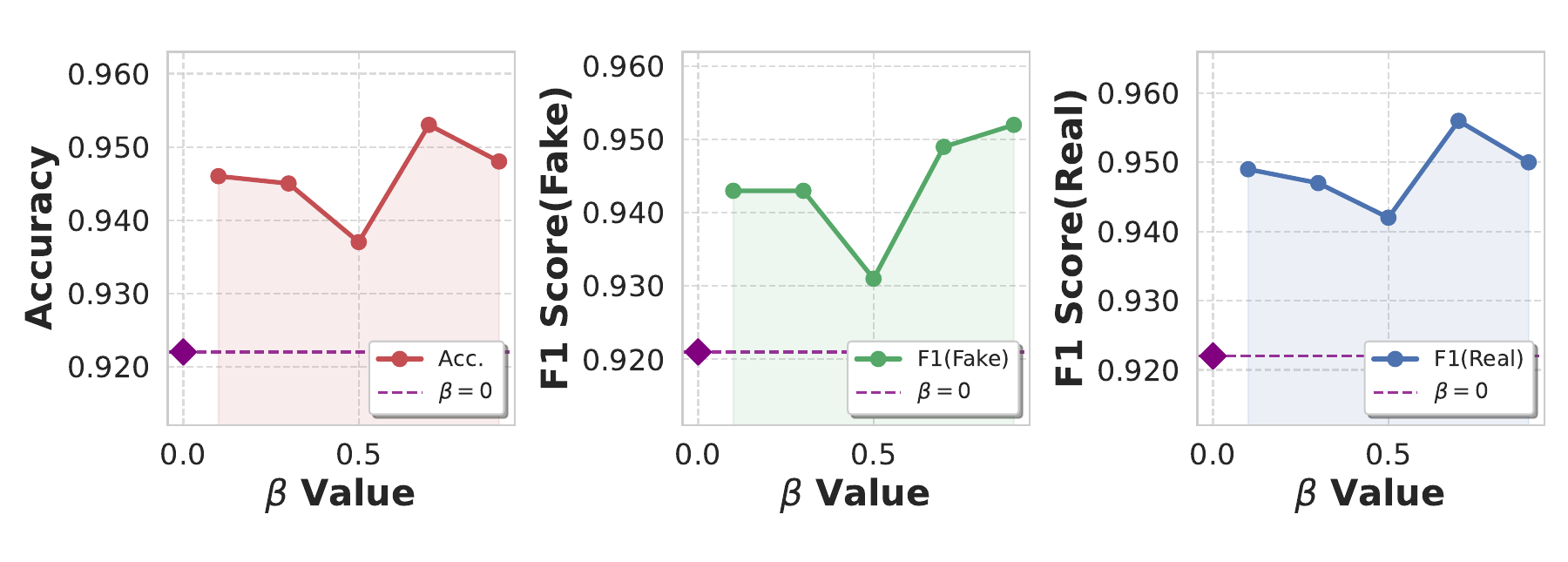}
    \end{subfigure}
    \vspace{-8pt}
    \begin{subfigure}{\columnwidth}
    \centering
    \includegraphics[width=\columnwidth]{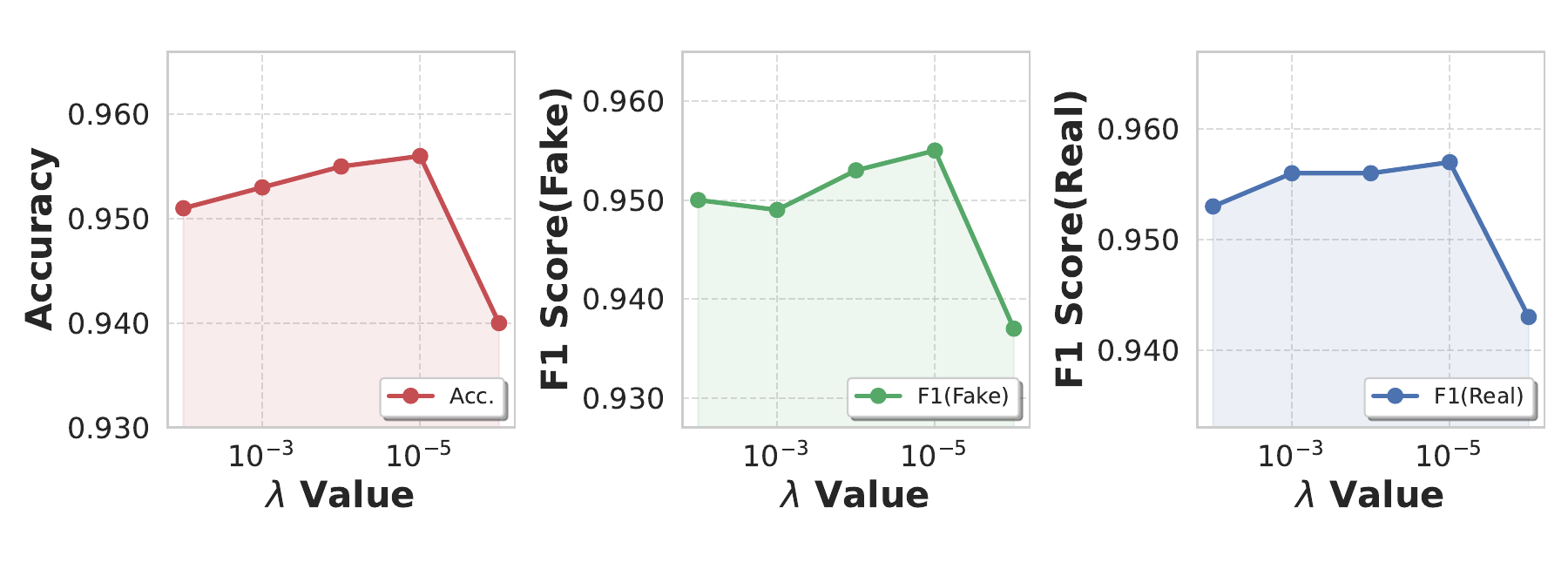}
    \end{subfigure}
    \vspace{-5pt}
    \caption{Performance metrics (accuracy, fake news F1 score, real news F1 score) for Weibo-21 dataset with different $\beta$ values and $\lambda$ values. Additional parameter analysis (Appendix~\ref{parameter_analysis_app}) for Weibo and GossipCop are shown in Figure~\ref{fig:beta_analysis_app} and Figure~\ref{fig:lambda_analysis_app}.}
    \label{fig:parameter_analysis}
\end{figure}

In addition, with only unimodal prediction agreement or only semantic alignment supervision, the model’s performance drops below that of the text-only ablation baseline, suggesting insufficient fusion of text and image modalities. Even without any modality interaction supervisions, the model still manages to learn some fusion patterns, resulting in improved performance over unimodal predictions. This indicates that the inherent architecture has some capability to integrate information across modalities in a less optimal manner. Notably, the best overall performance is achieved when both unimodal prediction agreement and semantic alignment supervision are combined. We conclude that, the joint supervision of unimodal prediction agreement and semantic alignment is essential for effectively capturing the interactions between image and text modalities, which maximizes the synergy between modalities, leading to an improved prediction performance.

\begin{figure*}[t]
    \setlength{\belowcaptionskip}{-5pt}
    \centering
    \begin{subfigure}{0.23\textwidth}
        \begin{tcolorbox}[colframe=gray, colback=white, boxrule=0.5pt, left=2pt, right=2pt, top=2pt, bottom=2pt, boxsep=1pt, arc=0mm, outer arc=0mm]
        \includegraphics[width=\linewidth, height=80pt]{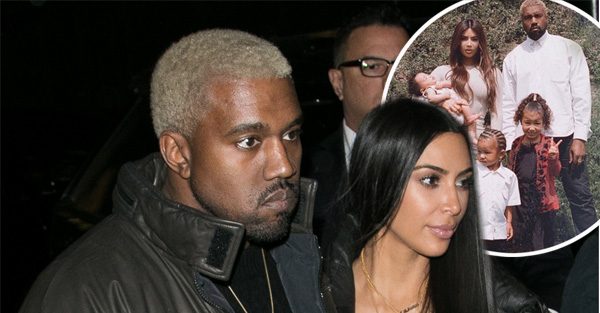}
        \parbox[t]{\textwidth}{\small Kanye West and wife Kim Kardashian West rarely smile. He told High Snobiety: ``People, you know, the paparazzi, always come up to me: 'Why you not smiling?' And I think, not smiling makes me..."}
        \end{tcolorbox}
        \parbox[t]{\textwidth}{\small \textbf{Label: $\boldsymbol{y=1}$; Predictions:}}
        \parbox[t]{\textwidth}{\small $\boldsymbol{\hat{y} = 0.73 | \hat{y}_\text{text} = 0.37 |  \hat{y}_\text{image} = 0.24}$ \rule{\textwidth}{0.4pt}}
        \includegraphics[width=\linewidth]{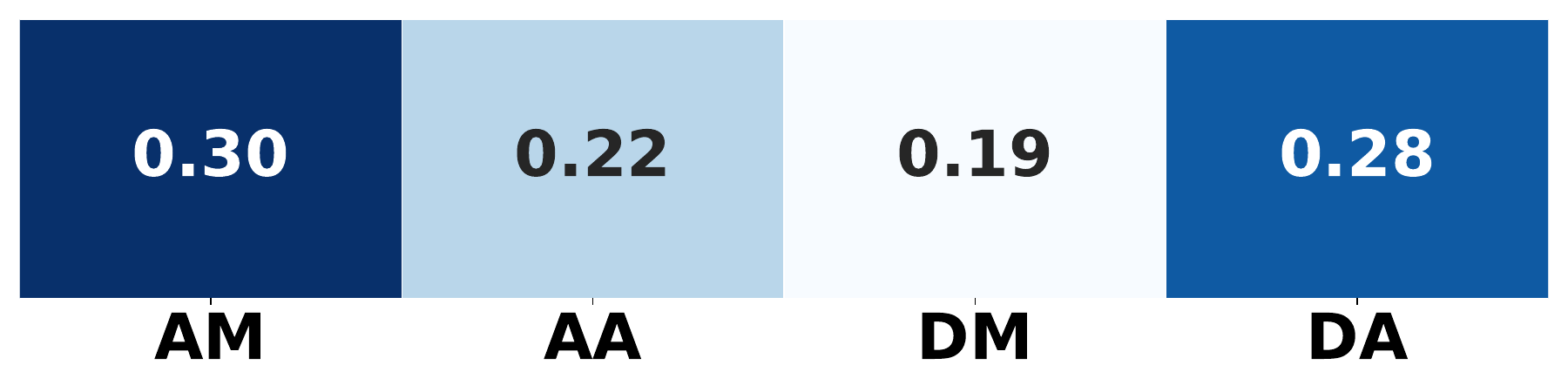}
        \caption{Agreed Misalignment}
        \label{fig:AM_case}
    \end{subfigure}
    \hfill
    \begin{subfigure}{0.23\textwidth}
        \begin{tcolorbox}[colframe=gray, colback=white, boxrule=0.5pt, left=2pt, right=2pt, top=2pt, bottom=2pt, boxsep=1pt, arc=0mm, outer arc=0mm]
        \includegraphics[width=\linewidth, height=80pt]{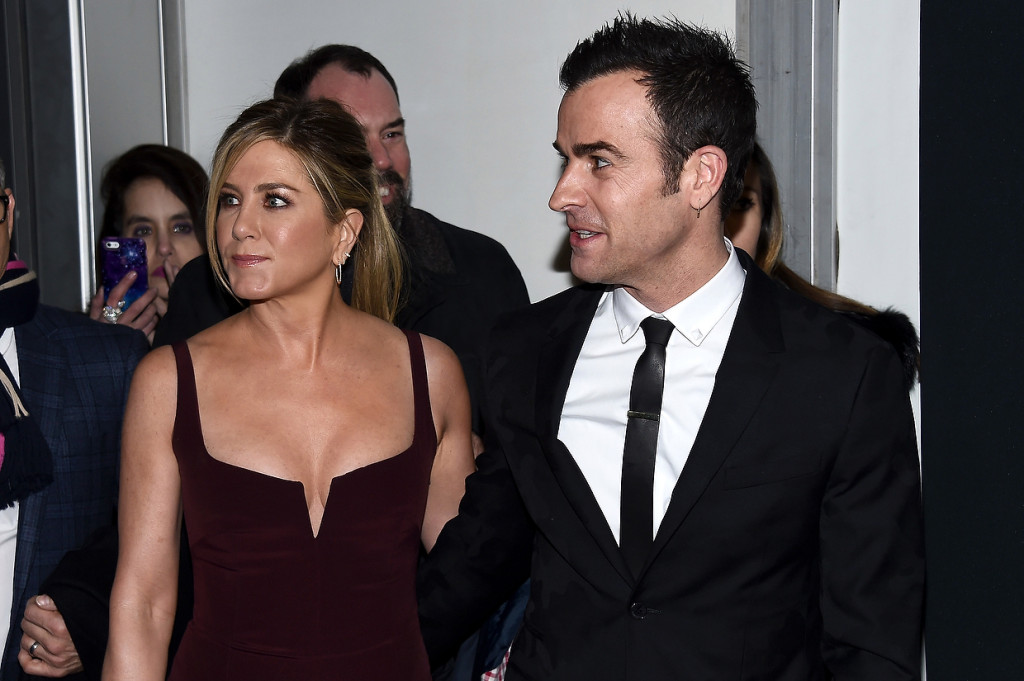}
        \parbox[t]{\textwidth}{\small Justin Theroux’s friends never believed he and Jennifer Aniston would last While the love story of Jennifer Aniston and Justin Theroux had its fans, sources close to the couple questioned whether it...}
        \end{tcolorbox}
        \parbox[t]{\textwidth}{\small \textbf{Label: $\boldsymbol{y=1}$; Predictions:}}
        \parbox[t]{\textwidth}{\small $\boldsymbol{\hat{y} = 0.99 | \hat{y}_\text{text} = 0.98 |  \hat{y}_\text{image} = 0.58}$} \rule{\textwidth}{0.4pt}
        \includegraphics[width=\linewidth]{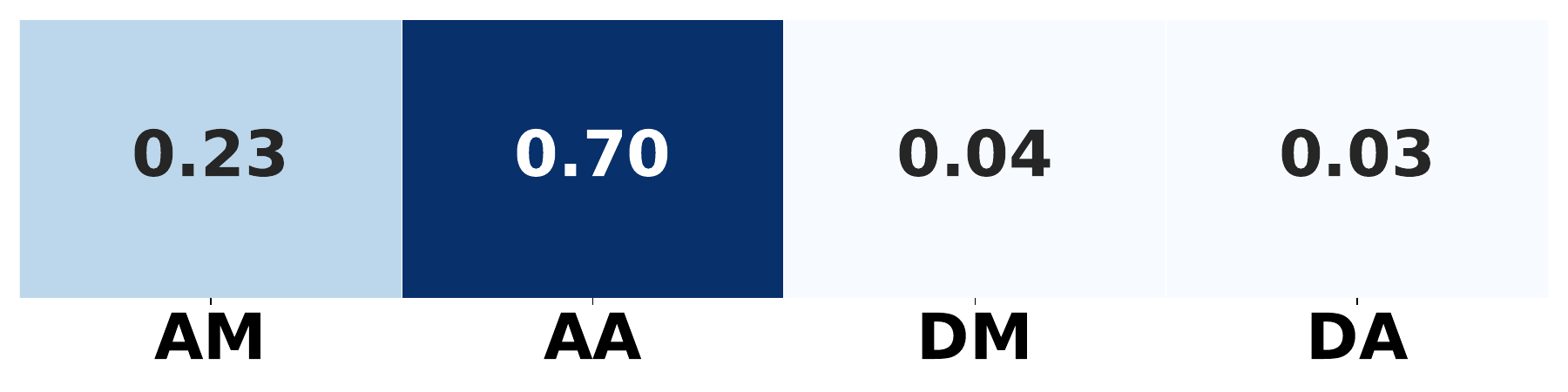}
        \caption{Agreed Alignment}
        \label{fig:AA_case}
    \end{subfigure}
    \hfill
    \begin{subfigure}{0.23\textwidth}
        \begin{tcolorbox}[colframe=gray, colback=white, boxrule=0.5pt, left=2pt, right=2pt, top=2pt, bottom=2pt, boxsep=1pt, arc=0mm, outer arc=0mm]
        \includegraphics[width=\linewidth, height=80pt]{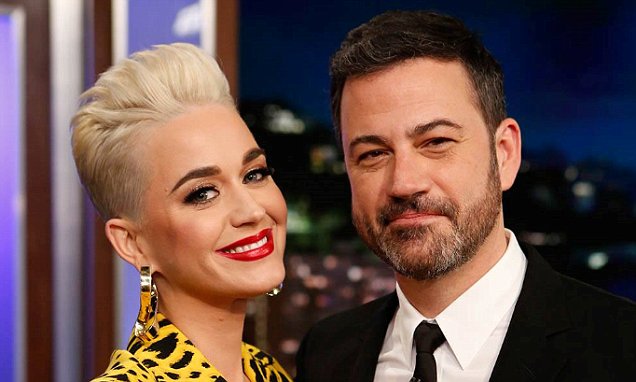}
        \parbox[t]{\textwidth}{\small Katy Perry slams Lionel Richie on Jimmy Kimmel Live! It appears there's drama at the American Idol judges table already. Katy Perry, who will be joined by Luke Bryant and Lionel Richie in...}
        \end{tcolorbox}
        \parbox[t]{\textwidth}{\small \textbf{Label: $\boldsymbol{y=1}$; Predictions:}}
            \parbox[t]{\textwidth}{\small $\boldsymbol{\hat{y} = 0.56 | \hat{y}_\text{text} = 0.32 |  \hat{y}_\text{image} = 0.50}$} 
        \rule{\textwidth}{0.4pt}
        \includegraphics[width=\linewidth]{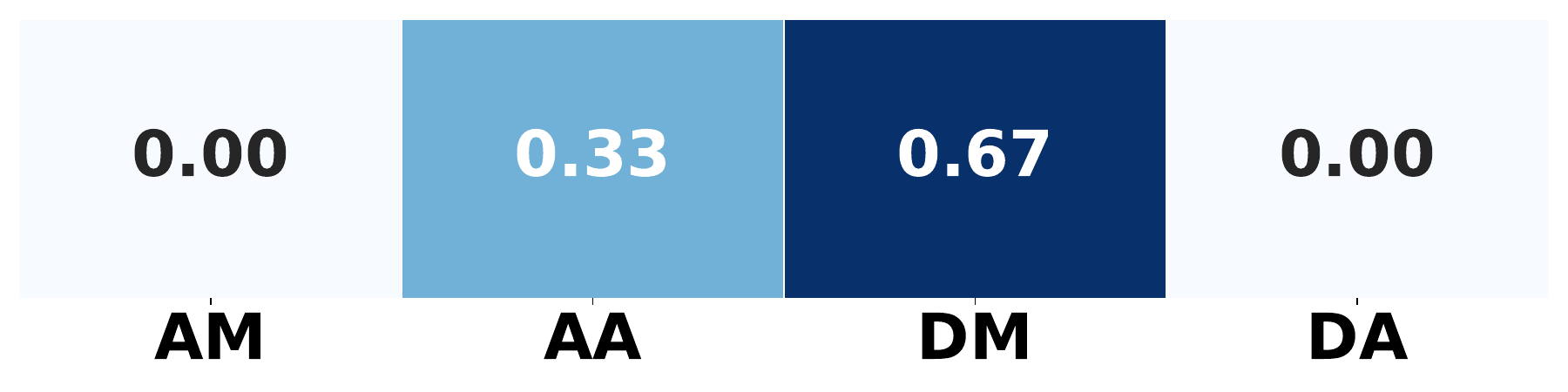}
        \caption{Disagreed Misalignment}
        \label{fig:DM_case}
    \end{subfigure} 
    \hfill
    \begin{subfigure}{0.23\textwidth}
        \begin{tcolorbox}[colframe=gray, colback=white, boxrule=0.5pt, left=2pt, right=2pt, top=2pt, bottom=2pt, boxsep=1pt, arc=0mm, outer arc=0mm]
        \includegraphics[width=\linewidth, height=80pt]{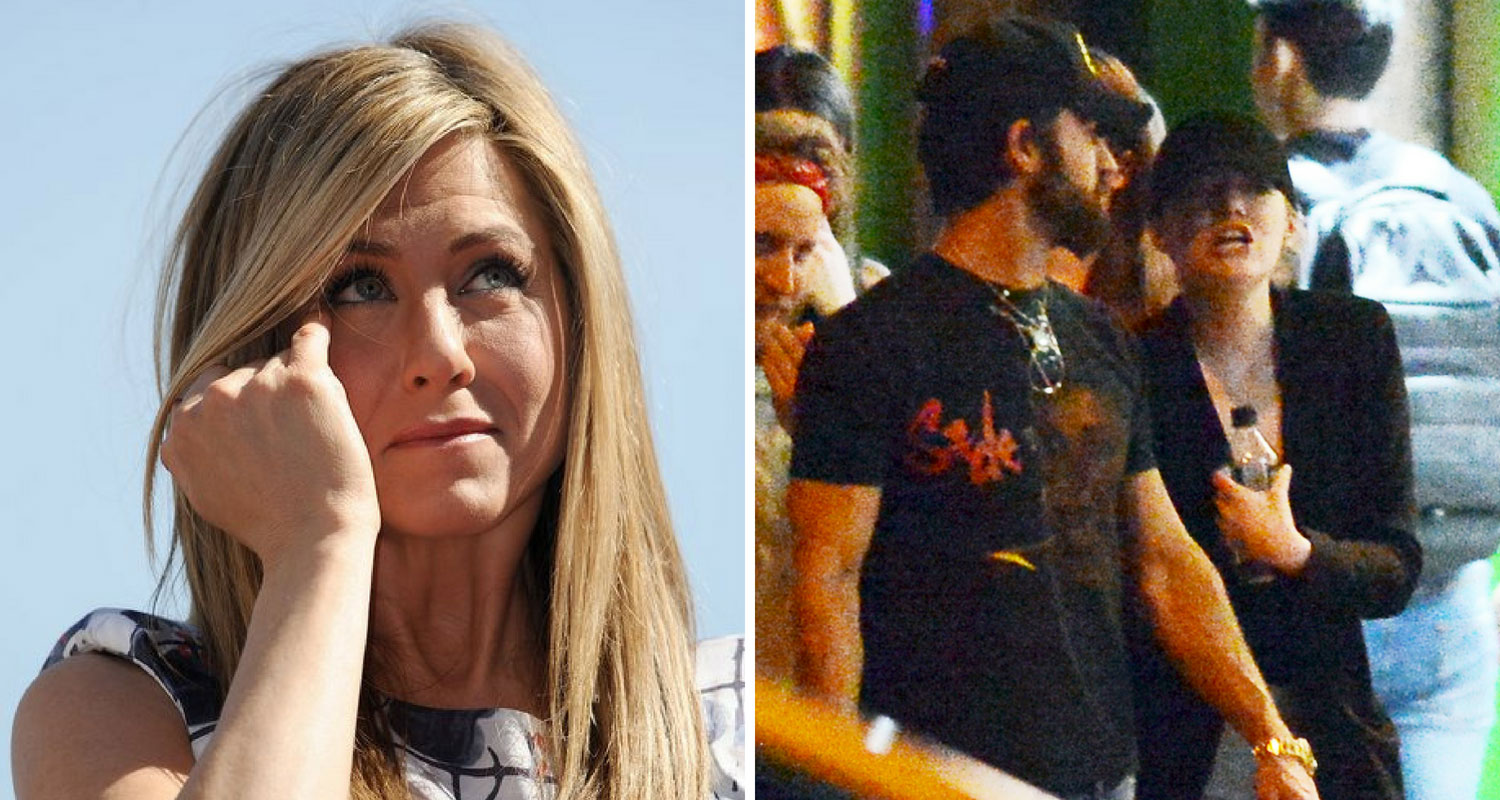}
        \parbox[t]{\textwidth}{\small Jen's fury as Justin commits the ultimate betrayal The pair became close friends on set when they filmed Netflix show Maniac and Jen is said to be wondering whether they have been dating...}
        \end{tcolorbox}
        \parbox[t]{\textwidth}{\small \textbf{Label: $\boldsymbol{y=1}$; Predictions:}}
        \parbox[t]{\textwidth}{\small $\boldsymbol{\hat{y} = 1.00 |  \hat{y}_\text{image} = 0.42 |\hat{y}_\text{text} = 0.98}$ \rule{\textwidth}{0.4pt}}
        \includegraphics[width=\linewidth]{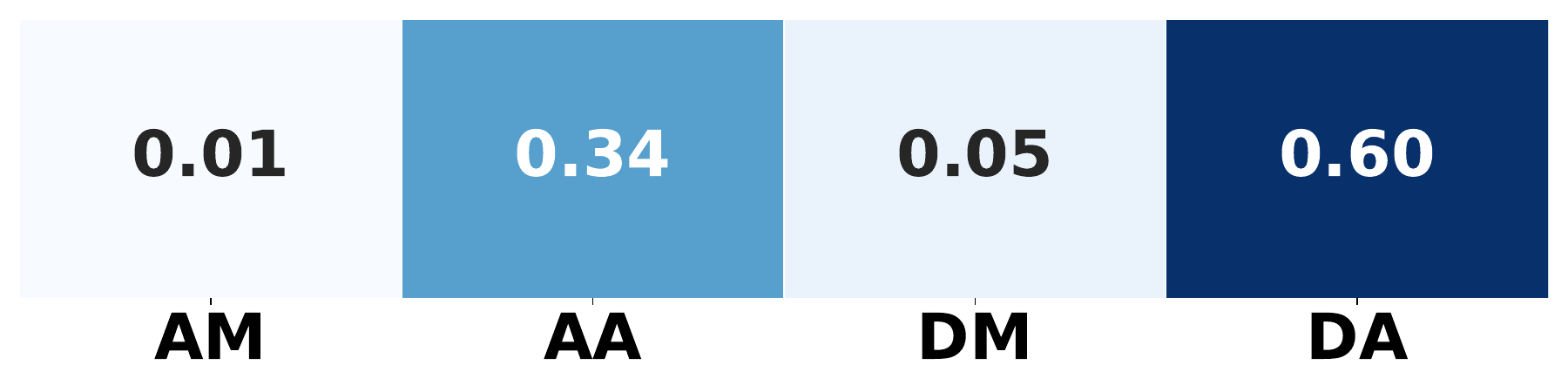}
        \caption{Disagreed Alignment}
        \label{fig:DA_case}
    \end{subfigure}
    \caption{Case study illustrating four instances dispatched by the model into their respective fusion experts: Agreed Misalignment (AM), Agreed Alignment (AA), Disagreed Misalignment (DM), and Disagreed Alignment (DA). Each example shows the model's final prediction ($\hat{y}$), unimodal predictions ($\hat{y}_{\text{text}}$, $\hat{y}_{\text{image}}$), and the dispatch vector for modality interactions (AM, AA, DM, DA). The examples demonstrate the model's capability to effectively address modality interaction-specific challenges.}
    \label{fig:case_study}
\end{figure*}

\vspace{-0.3cm}
\subsection{Parameter Analysis}
\label{parameter_analysis}

In this section, we analyze the sensitivity of the training process for modality interaction gating to two hyper-parameters: the modality interaction supervision weighting ($\beta$) and the learning rate ($\lambda$) for the interaction gating module. The parameter $\beta$ determines the contribution of modality interaction supervision in the final loss function, while $\lambda$ controls the update speed of the interaction gating module. For each parameter value, all other hyper-parameters are fixed according to the best configuration detailed in Appendix~\ref{implementation details}. Additional parameter analysis is presented in Appendix~\ref{parameter_analysis_app}.

As shown in Figure~\ref{fig:parameter_analysis}, we train MIMoE-FND using with different modality interaction loss weights $\beta \in \{0.1, 0.3, 0.5, 0.7, 0.9\}$ on all three datasets. Across all datasets, most models with nonzero $\beta$ outperform the baseline with no interaction supervision ($\beta=0$). The robustness to a wide range of $\beta$ values underscores the adaptability of MIMoE-FND and its ability to effectively leverage modality interaction supervision for enhanced fake news detection across all datasets. 
We also analyze the effect of varying the learning rate $\lambda \in \{10^{-2}, 10^{-3}, 10^{-4}, 10^{-5}, 10^{-6}\}$ for the interaction gating module, as presented in Figure~\ref{fig:parameter_analysis}. In Weibo-21, mid-range $\lambda$ values ($10^{-3}$ to $10^{-4}$) yield optimal performance, achieving the best trade-off between accuracy and F1 scores for both fake and real news. Conversely, extreme $\lambda$ values ($10^{-7}$ and $10^{-1}$) result in performance degradation, likely due to insufficient or overly aggressive updates to the interaction gating network.

\subsection{Gating Mechanism Case Study}
To analyze the behavior of the modality interaction gating mechanism under modality interaction supervision and task supervision, we conduct case studies using an English dataset to ensure broader relevance. Specifically, we select four instances, each dispatched to one of the fusion experts, illustrating the four modality interaction scenarios. In Agreed Misalignment (AM; Figure~\ref{fig:AM_case}), both modalities align while providing complementary and unique contributions. Due to semantic misalignment, the combined prediction ($\hat{y}=0.73$) is reversed compared to the individual predictions, highlighting that the fact that the AM expert synthesizes complementary cues from both modalities, leveraging synergistic information to address semantic misalignment and make the correct prediction. In Agreed Alignment (AA; Figure~\ref{fig:AA_case}), both modalities produce aligned predictions with high semantic alignment, as indicated by a dispatch weight of 0.7. The final prediction closely mirrors the text-only prediction, demonstrating redundancy from both text and image. For Disagreed Misalignment (DM; Figure~\ref{fig:DM_case}), the DM expert effectively detects the fake news by reconciling the contradictory signals even with low semantic alignment. 
Finally, in Disagreed Alignment (DA; Figure~\ref{fig:DA_case}), semantic information aligns despite unimodal prediction disagreements, resulting in a synergistic prediction that achieves a higher confidence score than either modality alone. Here the DA expert is able to examine the conflict between image and text predictions when they are semantically aligned and derives the final detection based on semantic-level reasoning. These case studies demonstrate that MIMoE-FND effectively utilizes modality-specific contributions and their interplay, showcasing robust multimodal inference capabilities.

%% file: sections/discussion.tex
\section{Conclusion}

In this paper, we introduce MIMoE-FND, a novel hierarchical MoE framework for multimodal fake news detection. Our approach models modality interactions through unimodal prediction agreement and semantic alignment. MIMoE-FND employs unimodal prediction divergence and CLIP-based semantic evaluation to supervise an interaction gating network that dynamically routes data instances to specialized fusion experts for precise multimodal fusion, where these experts leverage a modality-wise attention mechanism within an adapted MoE model to selectively attend to key modality features. 
Experimental results on three widely used multimodal FND datasets demonstrate that MIMoE-FND achieves superior performance compared to prior methods.

%% file: sections/appendix.tex
\begin{figure}[t]
    \setlength{\abovecaptionskip}{2pt}
    \centering
    \begin{subfigure}{\columnwidth}
        \centering
        \includegraphics[width=\columnwidth]{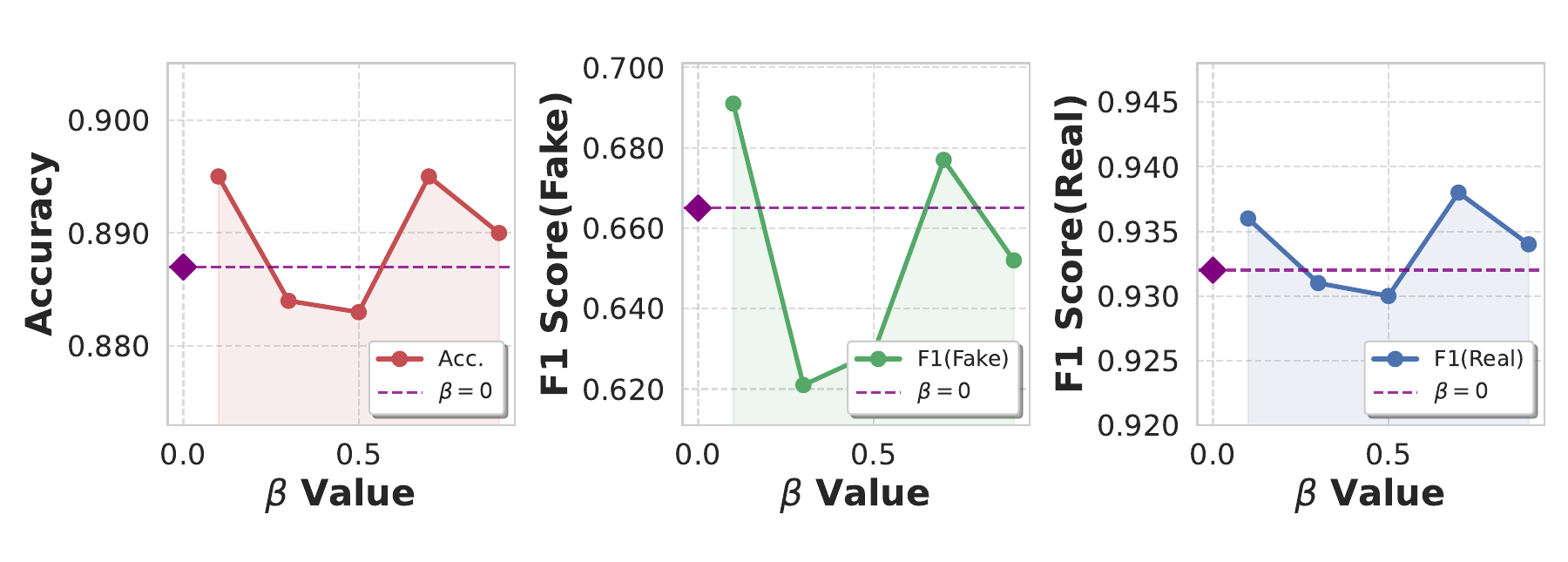}
        \captionsetup{skip=0pt}
        \caption{GossipCop}
        \label{fig:gossipcop_beta}
    \end{subfigure}
    \begin{subfigure}{\columnwidth}
        \centering
        \includegraphics[width=\columnwidth]{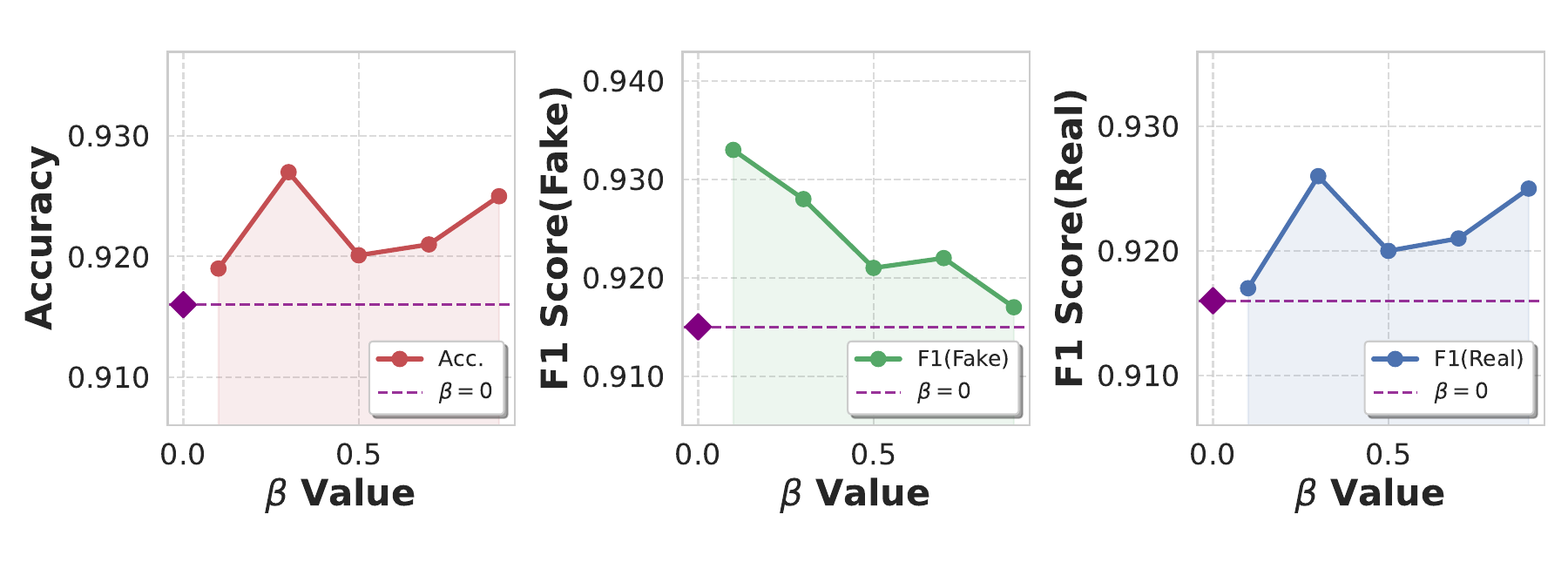}
        \captionsetup{skip=0pt}
        \caption{Weibo}
        \label{fig:weibo_beta}
    \end{subfigure}
    \caption{Performance metrics (accuracy, fake news F1 score, real news F1 score) for the GossipCop and Weibo datasets with different $\beta$ values. We notice that with a coarse $\beta$ parameter searching between 0 and 1, MIMoE-FND can significantly outperform ablated baseline ($\beta = 0$).}
    \label{fig:beta_analysis_app}
\end{figure}

\section{Experiment Setup}
\label{experimental_setting}
\subsection{Dataset} 
Since our method focuses on the modality interaction modeling, we select datasets with sufficient number of training data that has images and texts both accessible. Specifically, we consider three multimodal FND benchmark datasets commonly used by prior works, which are Weibo~\cite{weibo}, Weibo-21~\cite{weibo21} and GossipCop~\cite{gossipcop}. All three datasets are collected from social media with binary labels indicating the veracity of the multimodal news. Both Weibo and Weibo-21 are Chinese datasets containing image and text pairs collected from social media platform Weibo. Weibo has 3749 real news and 3783 fake news in training set, 1000 fake news and 996 real news in the test set. Weibo-21 was created in 2021, where more recent social posts were collected with 4640 real news and 4487 news in total. CossipCop is a dataset collected from fact verification website, which contains 7974 real news and 2036 fake news for training and 2285 real news and 545 fake news for testing. For Weibo and GossipCop we keep the train-test splits provided by original datasets in our evaluation. For Weibo-21, where train-test split is not provided by original dataset, we keep the same train-test split at a ratio of 9:1 of BMR~\cite{bootstrap}.

\vspace{-0.3cm}
\subsection{Implementation Details} 
\label{implementation details}
In our implementation, we use the pretrained checkpoint 'mae-pretrain-vit-base'\footnote{\url{https://github.com/facebookresearch/mae}} as our image encoder. For the Chinese datasets (Weibo and Weibo-21), we use 'bert-base-chinese'\footnote{\url{https://huggingface.co/google-bert/bert-base-chinese}}. as our text encoder, and for GossipCop, we use 'bert-base-uncased'\footnote{\url{https://huggingface.co/google-bert/bert-base-uncased}}. For CLIP embedding extractions, we utilize 'clip-vit-base-patch16'\footnote{\url{https://huggingface.co/openai/clip-vit-base-patch16}} for GossipCop and 'chinese-clip-vit-base-patch16'\footnote{\url{https://huggingface.co/OFA-Sys/chinese-clip-vit-base-patch16}} for the Chinese datasets. All MLPs in our network consist of two linear layers with a Sigmoid Linear Unit (SiLU) non-linearity in between. For all iMoE modules, we use two experts. Each expert network is implemented with a single-layer transformer block as defined in ViT with 4 attention heads. We employ the AdamW optimizer with a learning rate of $10^{-5}$ for optimization. 

We follow the same data preprocessing pipeline as provided by the official open-sourced code of EANN~\cite{EANN} and BMR~\cite{bootstrap}. Specifically, images are resized to $(224, 224)$ to suit the pretrained models. The maximum token length for both image and text is set to 197. For CLIP score evaluation, we truncate the input to meet the token upper bound of the pretrained models. To ensure reproducibility, all our models are trained for 50 epochs with a batch size of 24 and fixed random seed of 2024, where we report the best test accuracy obtained. Our model is trained on a single Nvidia A40 GPU with a training session of 50 epochs taking around 5 hours. All datasets we use are publicly available. Our source code will be made available upon acceptance.
\begin{figure}[t]
    \setlength{\abovecaptionskip}{2pt}
    \centering
    \begin{subfigure}{\columnwidth}
        \centering
        \includegraphics[width=\columnwidth]{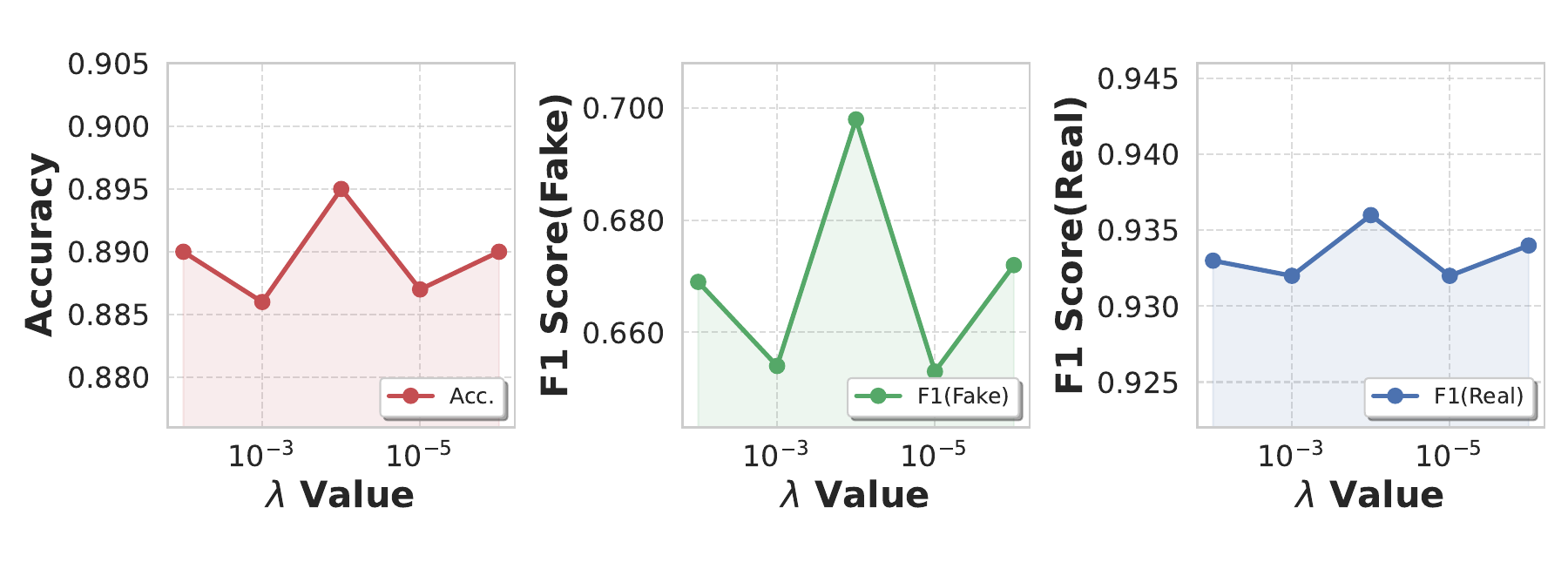}
        \captionsetup{skip=0pt}
        \caption{GossipCop}
        \label{fig:gossipcop_lambda}
    \end{subfigure}
    \begin{subfigure}{\columnwidth}
        \centering
        \includegraphics[width=\columnwidth]{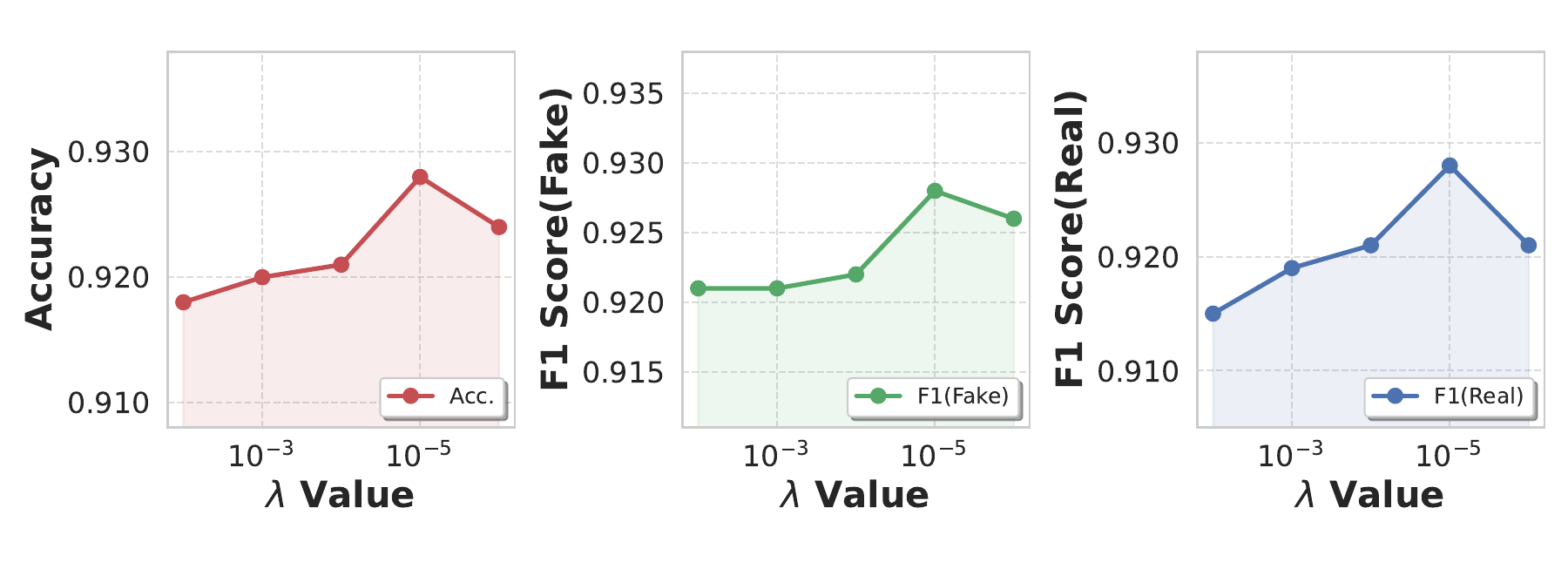}
        \captionsetup{skip=0pt}
        \caption{Weibo}
        \label{fig:weibo_lambda}
    \end{subfigure}
    \caption{Performance metrics (accuracy, fake news F1 score, real news F1 score) for the GossipCop and Weibo datasets with different $\lambda$ values.}
    \label{fig:lambda_analysis_app}
\end{figure}

\begin{figure}[h] 
    \centering
    \begin{minipage}[c]{0.05\textwidth}
        \centering
        \textbf{Weibo}
    \end{minipage}%
    \begin{minipage}[c]{0.45\textwidth}
        \centering
        \begin{subfigure}{0.3\linewidth}
            \includegraphics[width=\linewidth]{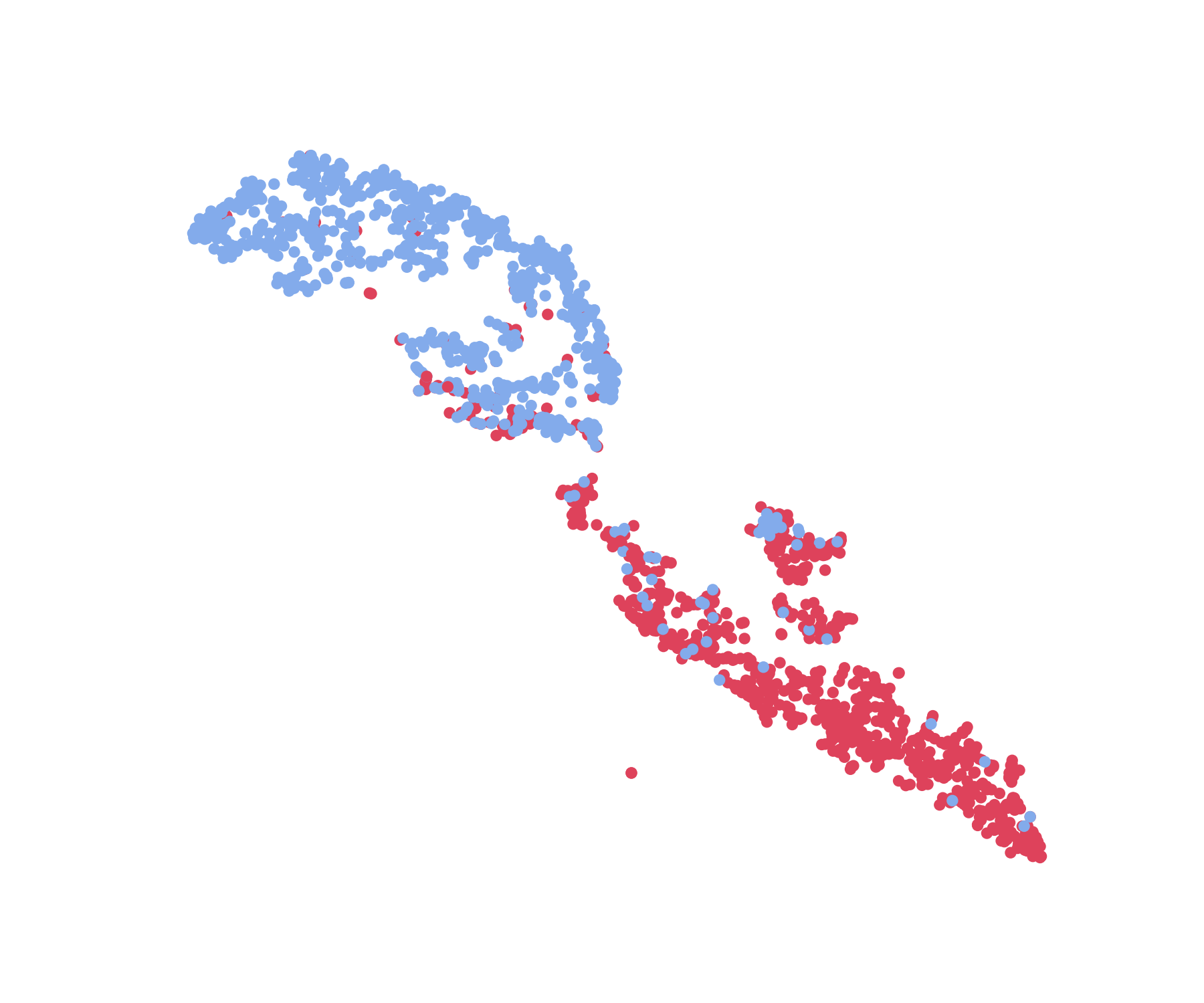}
        \end{subfigure}%
        \begin{subfigure}{0.3\linewidth}
            \includegraphics[width=\linewidth]{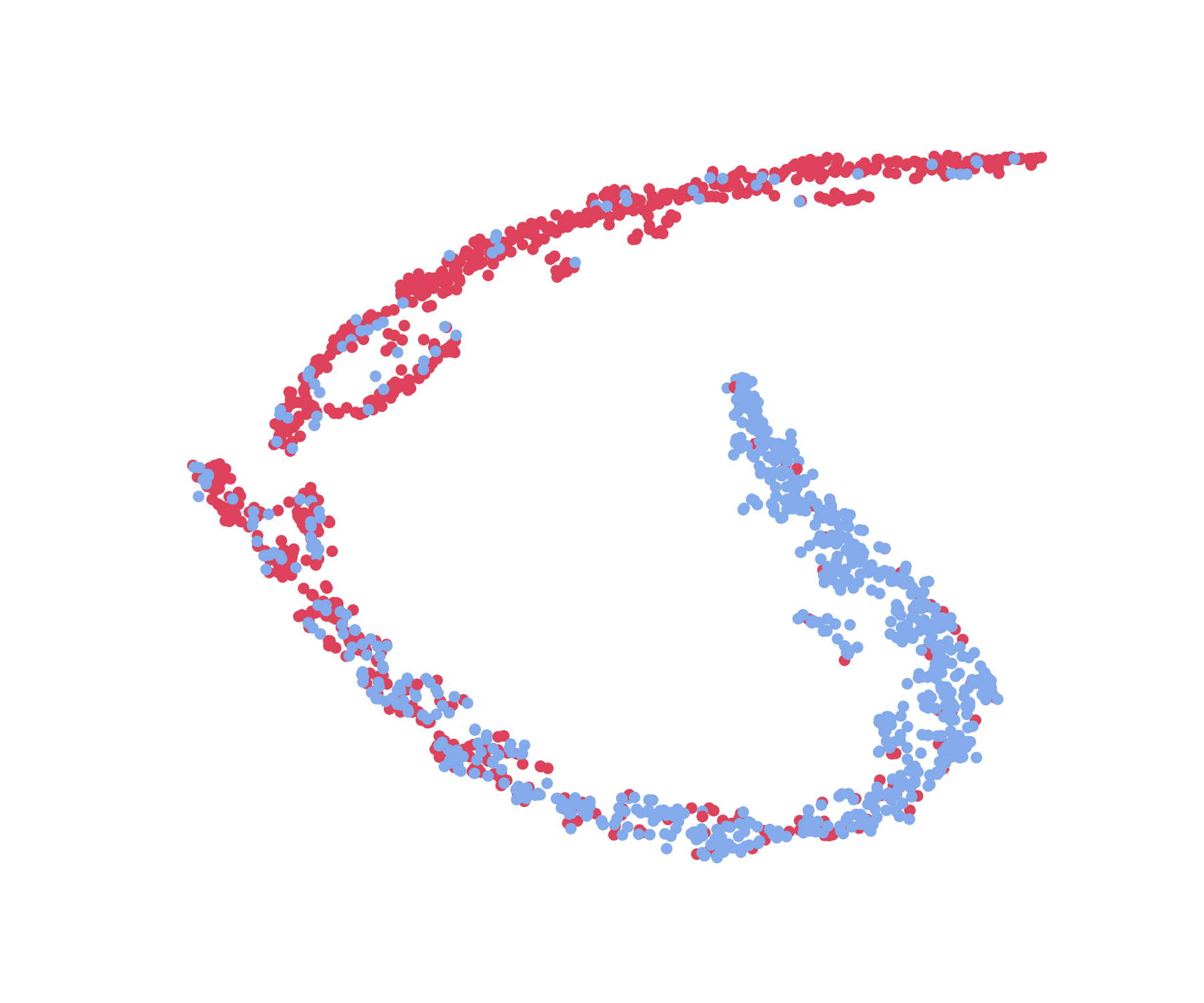}
        \end{subfigure}%
        \begin{subfigure}{0.3\linewidth}
            \includegraphics[width=\linewidth]{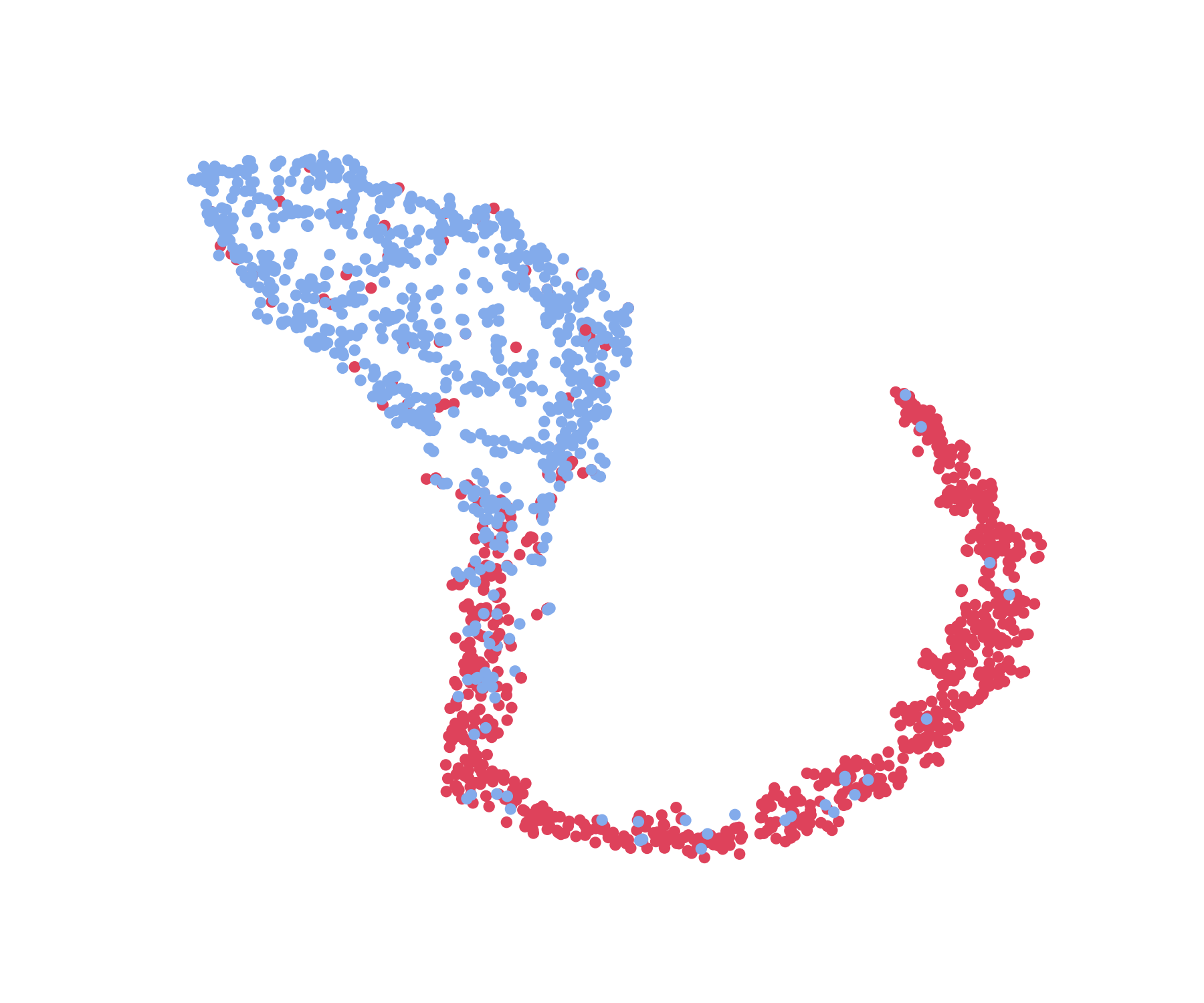}
        \end{subfigure}
    \end{minipage}

    \begin{minipage}[c]{0.05\textwidth}
        \centering
        \textbf{Gossip-Cop}
    \end{minipage}%
    \begin{minipage}[c]{0.45\textwidth}
        \centering
        \begin{subfigure}{0.3\linewidth}
            \includegraphics[width=\linewidth]{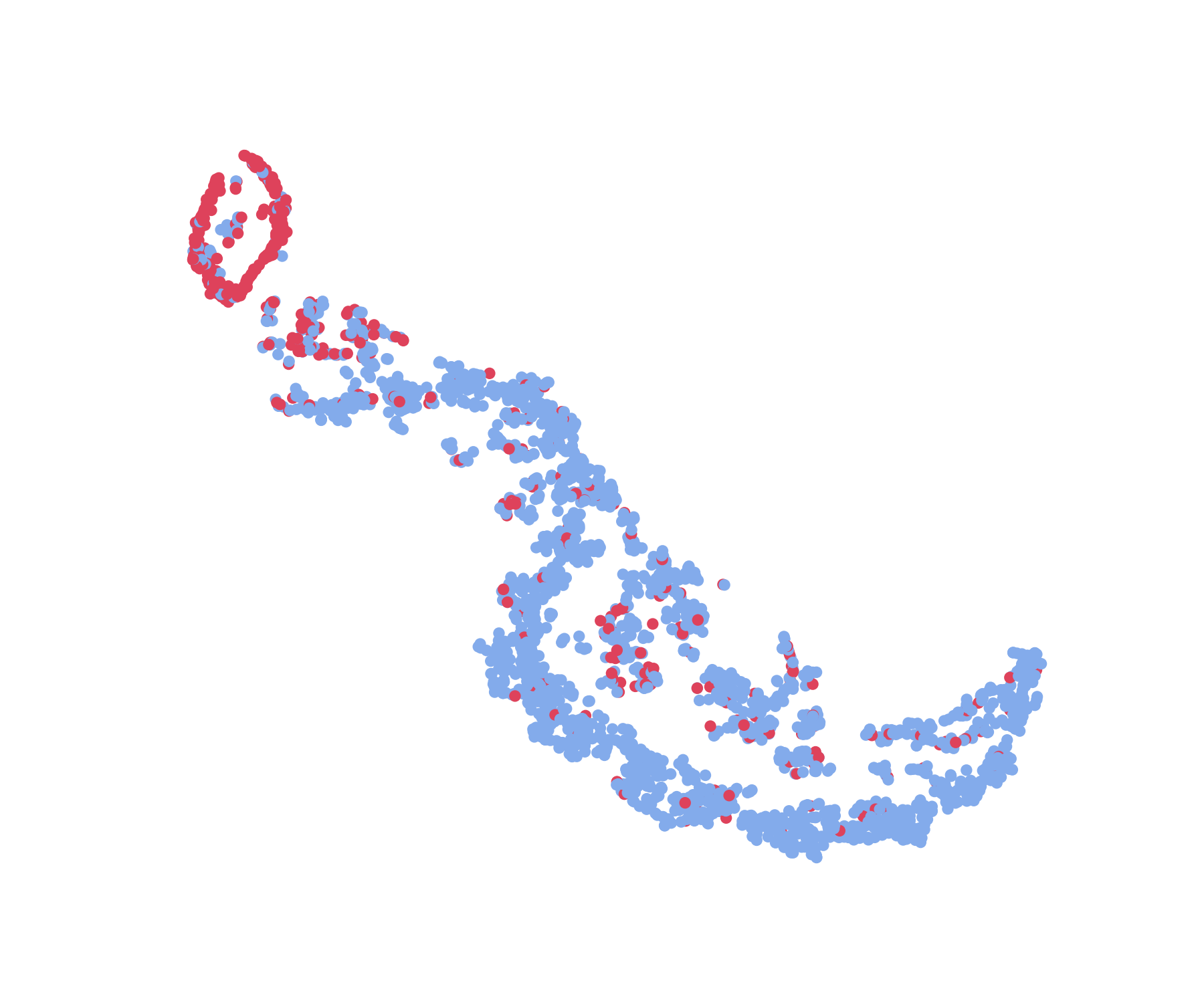}
        \end{subfigure}%
        \begin{subfigure}{0.3\linewidth}
            \includegraphics[width=\linewidth]{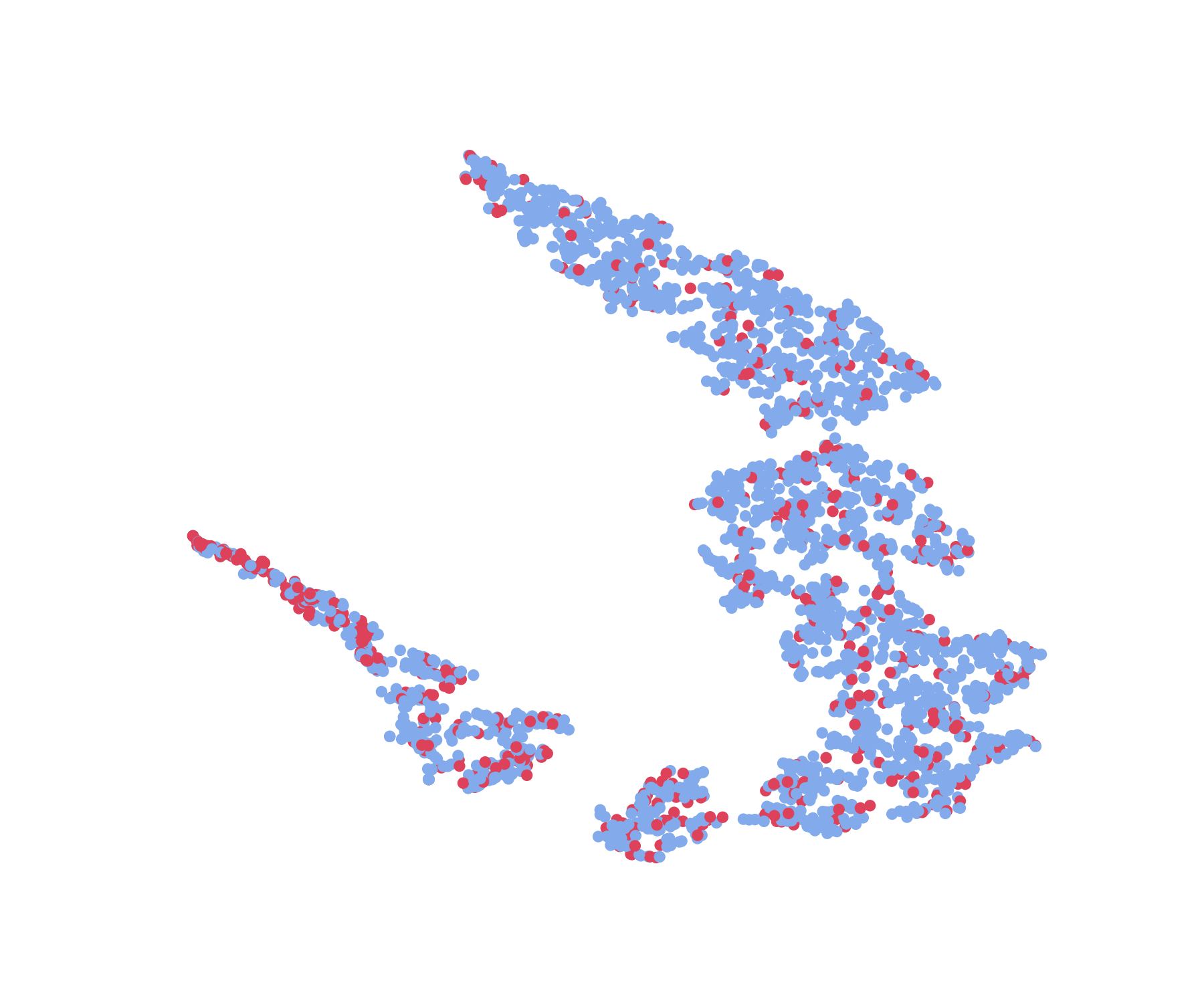}
        \end{subfigure}%
        \begin{subfigure}{0.3\linewidth}
            \includegraphics[width=\linewidth]{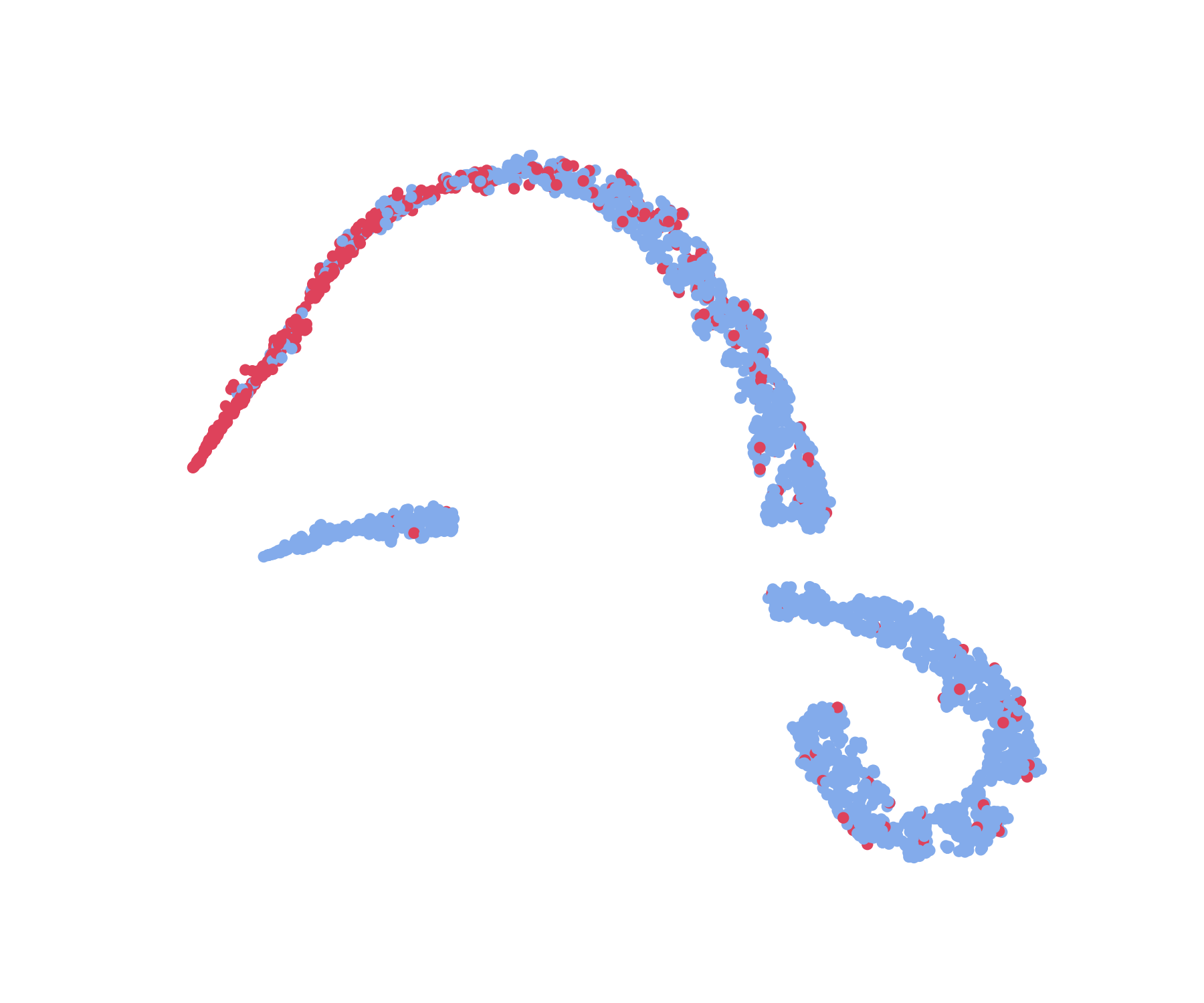}
        \end{subfigure}
    \end{minipage}

    \begin{minipage}[c]{0.05\textwidth}
        \centering
        \textbf{Weibo-21}
    \end{minipage}%
    \begin{minipage}[c]{0.45\textwidth}
        \centering
        \begin{subfigure}{0.3\linewidth}
            \includegraphics[width=\linewidth]{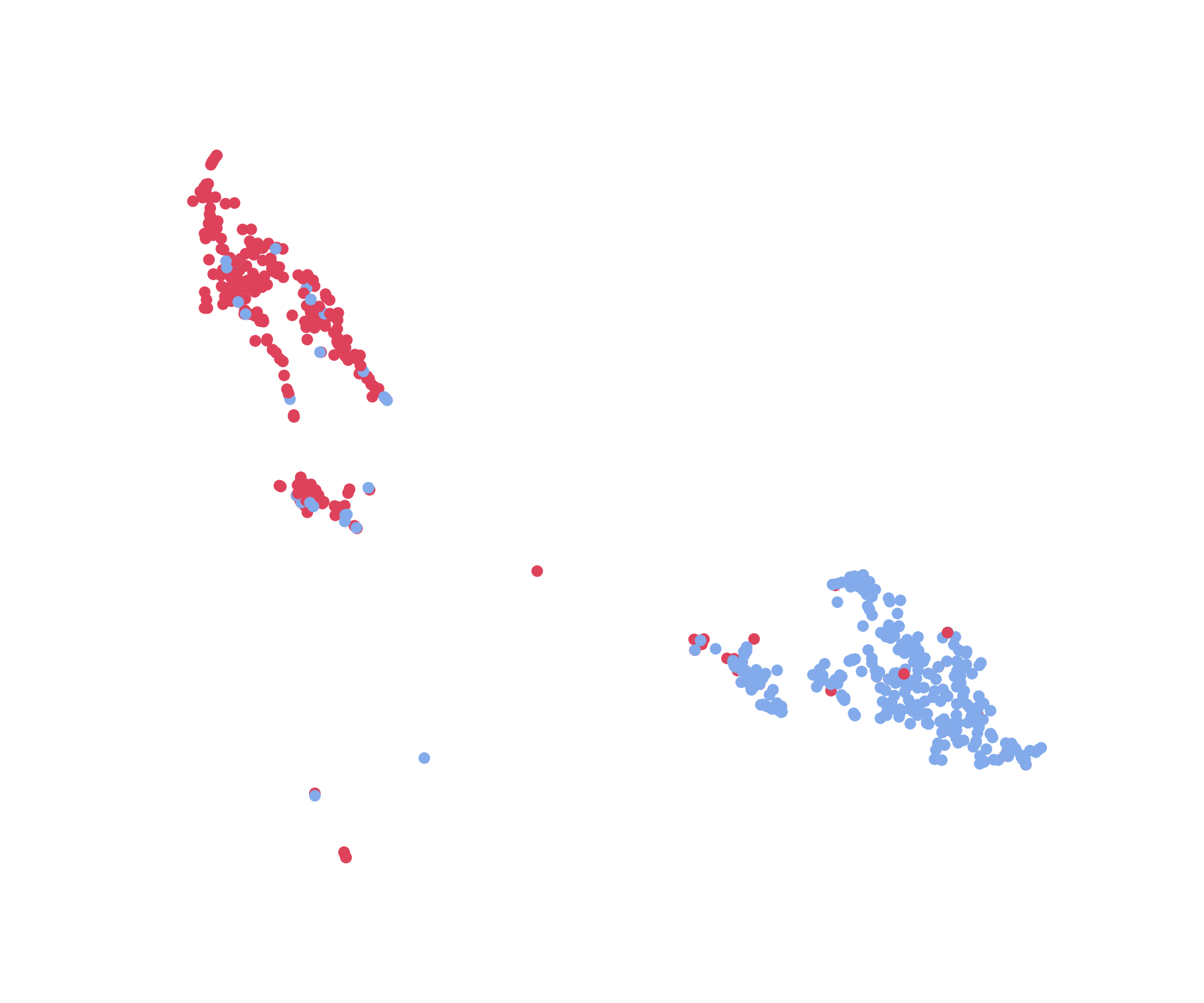}
            \caption*{MIMoE-FND}
        \end{subfigure}%
        \begin{subfigure}{0.3\linewidth}
            \includegraphics[width=\linewidth]{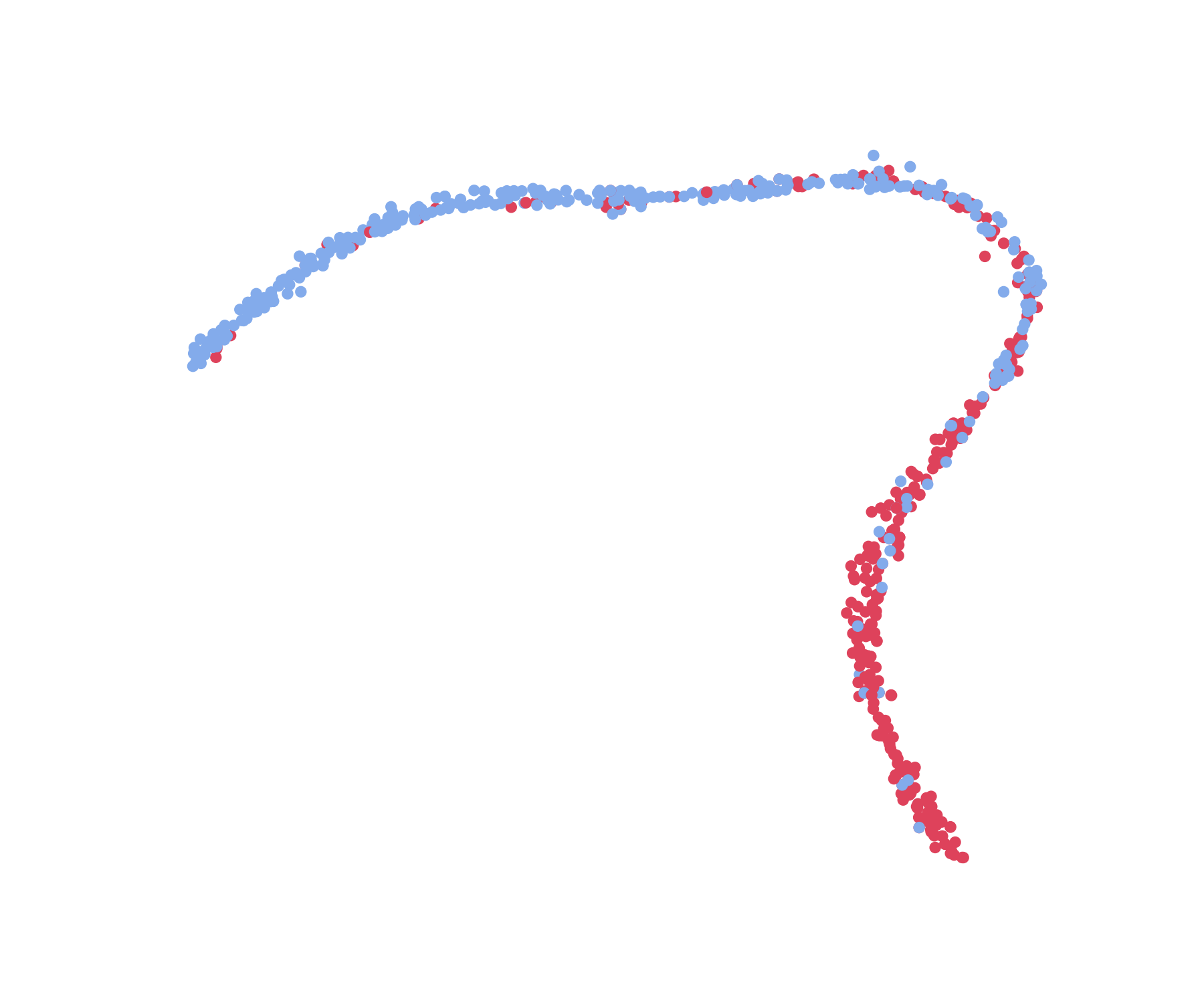}
            \caption*{Image-only}
        \end{subfigure}%
        \begin{subfigure}{0.3\linewidth}
            \includegraphics[width=\linewidth]{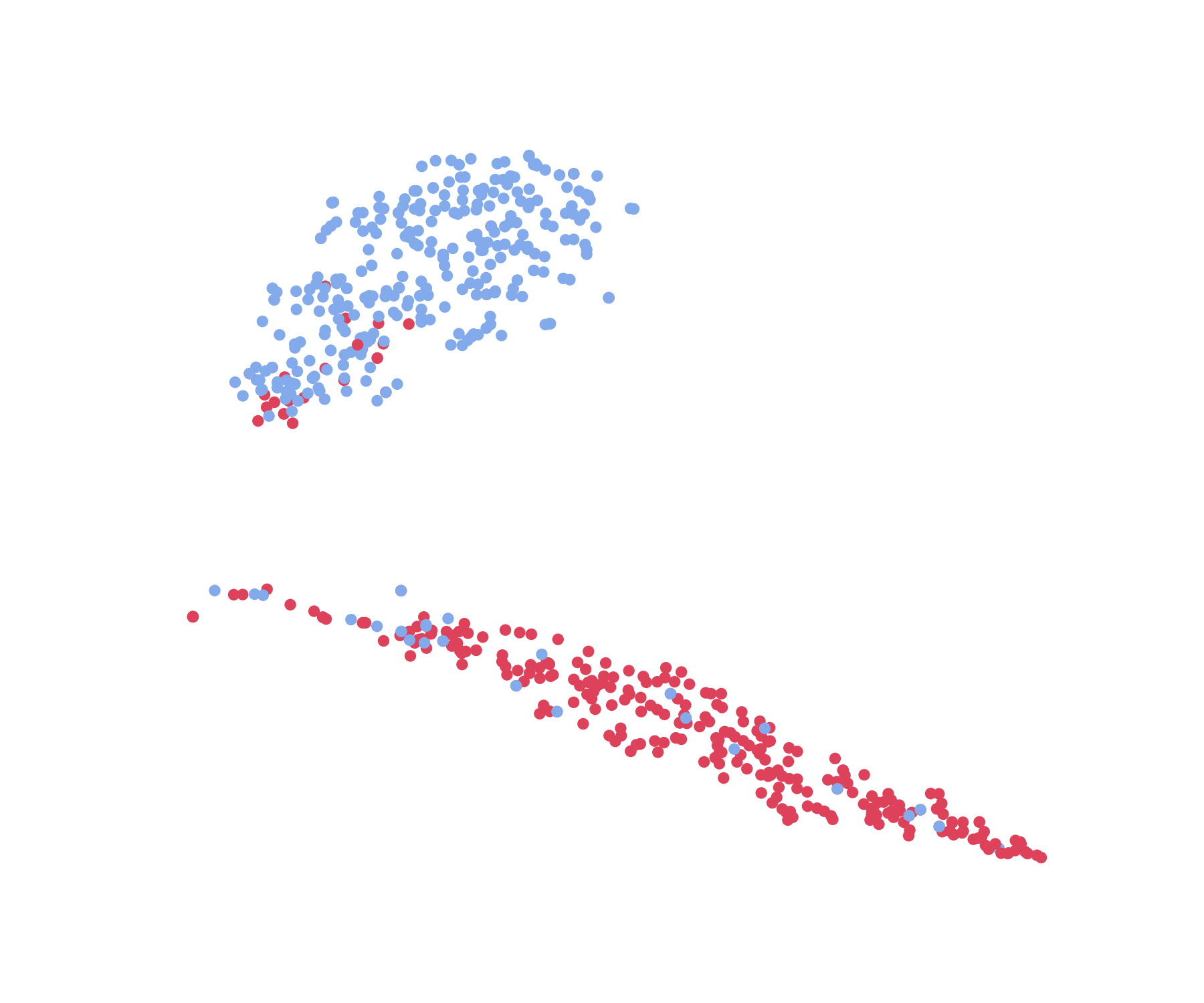}            \caption*{Text-only}
        \end{subfigure}
    \end{minipage}
    \vspace{-0.2cm}
    \caption{T-SNE visualizations of the features before final classifiers learned by our approach, image-only and text-only models.} 
    \label{fig:tsne}
\end{figure}

\vspace{-0.2cm}
\subsection{Baselines} We compare our method with a number of strong multimodal FND baselines, where we use their publicly available source code/pretrained model. We list the details of our baselines as follows:
\begin{itemize}
    \item \textbf{EANN}~\cite{EANN} is a multimodal FND solution that utilizes an event adversarial network to prevent model to overfit to a specific event. 
    \item \textbf{SAFE}~\cite{SAFE} is a method utilizing a similarity calculation between different modality representations to guide the multimodal FND. In our experiments, we adopt the official code released by the author. For the Chinese datasets which are not considered by the official implementation, we use the same text encoder as our approach and a pretrained ViT-GPT image captioning model\footnote{https://huggingface.co/yuanzhoulvpi/vit-gpt2-image-chinese-captioning}.
    \item \textbf{SpotFake}~\cite{spotfake} is a framework that consists of pretrained text and image encoders and a classifier to predict fake news with concatenated multimodal feature vectors.
    \item \textbf{CAFE}~\cite{CAFE} uses two VAEs to calculate cross-modal ambiguity in a probabilistic manner, which is then used to perform an adaptive fusion for multimodal FND.
    \item \textbf{BMR}~\cite{bootstrap} proposes to use a modified MMoE to refine and bootstrap multiview features for multimodal FND.
    \item \textbf{FND-CLIP}~\cite{fnd_clip} uses CLIP embeddings to bridge text and image features and guide an attention-based fusion module for more robust multimodal FND. For experiments on Chinese datasets, we adopt the same preprocessing as ours.
\end{itemize}

\section{Additional Parameter Analysis}
\label{parameter_analysis_app}
For $\beta$ values under the setting decribed in Section~\ref{parameter_analysis}, as shown in Figure~\ref{fig:parameter_analysis} and Figure~\ref{fig:beta_analysis_app}, the performance gains are more pronounced and consistent in Weibo and Weibo-21, where all tested $\beta$ values improve detection accuracy over the baseline. The consistent improvements in Weibo datasets indicates a stronger sensitivity to interaction supervision, which is likely due to the diversity of observed modality interactions in social media posts. For GossipCop, on the other hand, we observe two $\beta$ values where the performance drop below the ablated baseline. The performance drop of these $\beta$ values show that with lower diversity of modality interactions, partially enforcing modality interaction supervision can introduce routing instability, disrupting the model's ability to leverage the dominant unimodal signals effectively.
Overall, the robustness to a wide range of $\beta$ values underscores the adaptability of MIMoE-FND and its ability to effectively leverage modality interaction supervision for enhanced fake news detection across all datasets.

For $\lambda$ values, same as Weibo-21 results, mid-range $\lambda$ values ($10^{-3}$ to $10^{-4}$) yield optimal performance. Notably, the GossipCop dataset shows higher sensitivity to $\lambda$ variations compared to Weibo and Weibo-21, which is likely attributed to the imbalance in the number of samples across modality interactions.

\section{Qualitative Analysis}
In Figure~\ref{fig:tsne}, we extract feature vectors before classification heads for text-only, image-only and MIMoE-FND and visualize it using t-SNE visualizations. We observe that our method is able to achieve clear decision boundaries in all datasets. We also observe that compared to unimodal representations, MIMoE-FND features are more closely clustered, facilitating a more confident prediction.

\vspace{-0.2cm}
\section{Complexity Analysis}
\textbf{Complexity Analysis of iMoE Block: }First, the token attention vector $att_x$ is computed using a two-layer MLP, with complexity $O(N \cdot d^2)$ for an input sequence of shape $(N, d)$. The attention vector performs a weighted aggregation over $N$ tokens, adding $O(N \cdot d)$ complexity. The aggregated vector is then passed through another two-layer MLP to obtain the gate output, with complexity $O(d^2)$. Each of the $n_e$ expert networks applies a Vision Transformer block, with complexity $O(N^2 \cdot d + N \cdot d^2)$ per expert, resulting in $O(n_e \cdot (N^2 \cdot d + N \cdot d^2))$ for all experts. Finally, the gated the expert outputs aggregation adds a complexity of $O(n_e \cdot d)$.

Summing these terms, the overall complexity of the iMoE block is $O(N \cdot d^2 + d^2 + n_e \cdot (N^2 \cdot d + N \cdot d^2) + n_e \cdot d)$, which simplifies to $O(n_e \cdot (N^2 \cdot d + N \cdot d^2))$. This highlights dependencies on $N$ (number of tokens), $d$ (dimension of each token), and $n_e$ (number of experts), with quadratic dependence on $N$ due to the transformer block's attention mechanism.

\begin{table}[t]
\caption{Comparison of Processing Time and Standard Deviation for MIMoE-FND (Ours) and BMR}
\vspace{-0.2cm}
\label{tab:processing_time}
\resizebox{0.5\textwidth}{!}{
\begin{tabular}{@{}lccc@{}}
\toprule
\textbf{Language} & \textbf{Method} & \textbf{Mean Processing Time (s)} & \textbf{Standard Deviation (s)} \\
\midrule
\multirow{2}{*}{English} & Ours & 0.1456 & 0.0194 \\
                           & BMR  & 0.1138 & 0.0336 \\
\midrule
\multirow{2}{*}{Chinese}  & Ours & 0.1503 & 0.0139 \\
                           & BMR  & 0.1151 & 0.0259 \\
\bottomrule
\end{tabular}}
\end{table}
\textbf{Complexity Analysis of MIMoE-FND: }In our overall hierachical MoE design of MIMoE-FND, we adopt iMoE blocks for feature refinement and multimodal fusion. In feature refinement stage, each input modality is refined by an iMoE block, resulting in a time complexity of $O(M \cdot n_e \cdot (N^2 \cdot d + N \cdot d^2))$ for $M$ modalities. In fusion stage, to account for four modality interaction scenarios, we adopt four fusion experts, each of which is implemented as an iMoE block, enabling adaptive and context aware feature aggregation. Consider $k$ experts are sparsely activated, the fusion stage complexity is then $O(k \cdot n_e \cdot (N^2 \cdot d + N \cdot d^2))$, where $k \le 4$. Overall, this yields a time complexity of $O((k+M) \cdot n_e \cdot (N^2 \cdot d + N \cdot d^2))$. Empirically, the modality channel number $M$ and fusion expert activation number $k$ are fixed as small constants. The overall time complexity is dominated by the sequence length, which is a common scalability bottleneck from transformer architecture shared by the majority of recent approaches. In Table~\ref{tab:processing_time}, we measure the 16-sample batch inference time for both languages and compare it to the previous state-of-the-art solution, BMR. We observe that our method introduces a slight computational overhead ($\sim$0.03 seconds) attributable to its larger parameter size. However, this overhead is offset by the substantial performance improvements achieved, demonstrating the effectiveness of our approach.